\documentclass{article}

\usepackage[preprint]{neurips_2026}
\usepackage{amsmath}
\usepackage{tikz}
\usetikzlibrary{arrows.meta,positioning}
\usepackage{xcolor}
\usepackage{graphicx}
\usepackage{listings}
\usepackage{booktabs, makecell}

\lstdefinestyle{promptstyle}{
  basicstyle=\ttfamily\scriptsize,
  breaklines=true,
  breakatwhitespace=false,
  columns=fullflexible,
  keepspaces=true,
  frame=single,
  xleftmargin=0.5em,
  xrightmargin=0.5em,
  aboveskip=0.6em,
  belowskip=0.6em,
  literate={—}{{---}}1
}

\usepackage[utf8]{inputenc} 
\usepackage[T1]{fontenc}    
\usepackage{hyperref}       
\usepackage{url}            
\usepackage{booktabs}       
\usepackage{amsfonts}       
\usepackage{nicefrac}       
\usepackage{microtype}      
\usepackage{xcolor}         
\usepackage{algorithm}
\usepackage{algorithmic}
\usepackage{titlesec}
\hypersetup{hypertexnames=false}

\title{Ambig-DS: A Benchmark for Task-Framing Ambiguity in Data-Science Agents
}

\author{%
  Josefa Lia Stoisser$^{1}$ \And
  Marc Boubnovski Martell$^{1}$ \And
  Sidsel Boldsen$^{1}$ \And
  Kaspar M\"artens$^{1}$ \And
  Robert Kitchen$^{1}$ \\[3.5ex]
  $^{1}$Novo Nordisk \\
  London, UK
}

\begin{document}




\maketitle

\begin{abstract}
As data-science agents shift from co-pilots to auto-pilots, silent misframing becomes a critical failure mode.  Agents quietly commit to plausible but unintended task framings, producing clean, executable artifacts that hide their incorrect assessment of the task.  Existing benchmarks score whether the pipeline runs, ignoring whether the agent recognized the task was underspecified. We introduce Ambig-DS, two diagnostic suites — one for prediction-target ambiguity (Ambig-DS-Target, 51 tasks built on DSBench, a tabular modeling benchmark) and one for evaluation-objective ambiguity (Ambig-DS-Objective, 61 tasks built on MLE-bench, a Kaggle-style ML competition benchmark) — constructed so that scoring uses each source benchmark's original evaluator. For every task we pair the original, fully specified version with an ambiguous variant produced by controlled edits; a human-and-LLM verification pipeline confirms each variant admits multiple plausible interpretations with decision-relevant consequences. The suites are analyzed independently and ambiguity lowers performance in both. Across five agents spanning efficient to frontier-class models, we find in our controlled diagnostic setting: (i) failures are silent commitments — wrong-target submissions on Target, wrong-metric or non-committal baseline submissions on Objective — rather than execution errors; (ii) allowing the agent to ask one clarifying question recovers much of the loss under idealized conditions, suggesting missing framing information drives a substantial part of the observed degradation; but (iii) agents cannot reliably tell when to use it: permissive prompts induce over-asking on clear tasks, while conservative prompts induce silent defaulting on ambiguous ones. Recognizing target and objective underspecification, not pipeline execution, is the bottleneck missing from standard DS-agent evaluations.

\end{abstract}

\section{Introduction}
\label{sec:intro}
The five frontier code-based agents we test, given a Kaggle-style prompt and the underlying CSV, commit to the wrong prediction target on $39$--$63\%$ of ambiguous tasks --- and do not flag the ambiguity in any of them. The pipeline executes, the model trains, the submission is well-formed, and the source-benchmark evaluator accepts the artifact as valid. We call this failure mode \emph{unflagged misframing}: a coherent, executable artifact that silently solves a plausible-but-unintended task, with no signal for a downstream consumer to catch (illustrated in Appendix Figure~\ref{fig:framing_ambiguity}).

Recent code-agent benchmarks~\citep{lai2023ds,huang2024code,huang2023mlagentbench,jing2024dsbench,chan2024mle,zhang2025datascibench,nathani2025mlgym,wijk2024re} measure pipeline execution under prompts that already fix the prediction target, output form, and evaluation metric. Real data-science work routinely lacks these commitments~\citep{passi2019problem}: stakeholders supply a goal and a dataset, and the modeler decides what counts as the target and what counts as success. Whether agents can recognize when those decisions are underdetermined --- and ask before acting --- is not measured by current benchmarks, and is the competence we evaluate.

We introduce \textbf{Ambig-DS}, a paired diagnostic for \emph{framing recognition}: the competence of noticing that the prompt--data observation does not justify a unique commitment, and asking before acting. Ambig-DS holds the source-benchmark evaluator fixed and intervenes only on the prompt--data pair. \emph{Ambig-DS-Target} (51 paired tasks on DSBench) isolates prediction-target ambiguity; \emph{Ambig-DS-Objective} (61 paired tasks on MLE-bench) isolates evaluation-metric ambiguity. The intended framing is used only for scoring; the desired behavior is not to recover it from an underdetermined observation, but to recognize the underdetermination and ask. Each ambiguous variant is human-reviewed against a four-item retention checklist (plausible alternatives, ambiguity preservation, decision relevance, task preservation), with cross-family LLM-verifier audits confirming the labels. We analyze each suite independently and do not pool results or compare effect magnitudes across suites.

Across five code-based agents, ambiguity induces unflagged misframing on $39$--$63\%$ of Ambig-DS-Target runs and silent failure on $16$--$62\%$ of Ambig-DS-Objective runs --- failures of framing recognition rather than execution. An idealized one-shot clarification oracle recovers much of the loss when invoked, but no model--prompt-policy combination we test achieves both calibrated asking on ambiguous tasks and restraint on fully-specified ones: agents can use missing framing information when offered but cannot reliably detect when to ask. Data-science agent evaluation should therefore test framing recognition before pipeline execution.

\paragraph{Contributions.} We (i) formalize task framing and isolate two of its core variables (prediction target and evaluation objective); (ii) release \textbf{Ambig-DS}, two independently analyzed diagnostic suites (51 DSBench + 61 MLE-bench paired tasks) built via controlled prompt--data interventions verified by humans and cross-family LLM verifiers; and (iii) quantify the resulting misframing and ask--act miscalibration across five frontier agents as above.

\section{Related Work}
\label{sec:related_work}

\paragraph{Data Science / ML Benchmarks.}
Recent benchmarks evaluate LLMs and agents on increasingly realistic data-science and ML workflows. DS-1000~\citep{lai2023ds} and DA-Code~\citep{huang2024code} focus on executable data-science code generation, while MLAgentBench~\citep{huang2023mlagentbench}, DSBench~\citep{jing2024dsbench}, MLE-bench~\citep{chan2024mle}, DataSciBench~\citep{zhang2025datascibench}, MLGym~\citep{nathani2025mlgym}, and RE-Bench~\citep{wijk2024re} evaluate broader agentic workflows involving data preparation, modeling, experimentation, and iterative improvement, and serve as substrates for recent code-DS agent systems~\citep{nam2025mle, schmidgall2025agent, stoisser2025query}. These benchmarks make DS/ML-agent evaluation more realistic, but still assume a specified task: the agent is told what to predict, optimize, and measure. Ambig-DS instead evaluates whether agents can recognize and resolve underspecified task framings before executing a pipeline, motivated by work showing that problem formulation choices are discretionary and consequential~\citep{passi2019problem}.

\paragraph{Ambiguity, Clarification, and Ask-Before-Act Agents.}
Recognizing underspecification\footnote{We use \emph{underspecification} in the task-framing sense: the prompt--data observation does not pin down a unique supervised-learning problem. This is distinct from the model-level sense of \citet{d2022underspecification}.} has been studied across a sequence of progressively more action-laden settings. In open-domain QA and information seeking, prior work measures and trains for clarification of ambiguous questions, information needs, and entity references~\citep{min2020ambigqa,kuhn2022clam,zhang2024clamber,lee2024ambigdocs,park2025mirage, stoisser2025towardsLabel}; in dialogue, work studies when and what to ask, and how clarification reshapes downstream answers~\citep{zhang2025clarify,zhang2024modeling,chen2024learning,zhao2026and,luo2025clarifymt}; for tool-using and planning agents, clarification is studied over unclear instructions, missing arguments, and uncertain tool calls~\citep{zhang2024ask,mitra2026recap,wang2025learning,zhang2025asktoact,suri2025structured,liu2024uncertainty,zhang2024toolbehonest, stoisser2025towards}. A common arc holds across these subfields: the community first \emph{measures} silent failure under ambiguity, then builds ask-before-act benchmarks, then trains for calibrated asking. For executable predictive data-science workflows --- where the unit of ambiguity is the supervised-learning framing rather than a question, a tool argument, or a plan step --- the first step is missing. Ambig-DS supplies it.

\paragraph{Executable and Data-Centric Underspecification.}
Closest to our setting are works where ambiguity affects executable artifacts. Code-generation and software-engineering work studies ambiguous requirements, clarification-aware training, and interactive clarification before implementation~\citep{mu2023clarifygpt,wu2025clarifycoder,li2023python,li2026clareval}, including on real software-engineering tickets~\citep{vijayvargiya2026ambig}, establishing ask--act evaluation as a mature setting for code and software agents. Data-centric work studies ambiguity in tabular analysis and natural-language table queries~\citep{li2025large,gomm2025we}, while text-to-SQL work considers cases where one natural-language query maps to multiple plausible SQL programs~\citep{bhaskar2023benchmarking, stoisser2025struct, stoisser2025sparks}. Across these settings, ambiguity affects the surface form of the artifact while the underlying task type is fixed; Ambig-DS targets ambiguity in the supervised-learning framing itself --- what target to predict or which objective to optimize --- so that misframing yields a fully valid pipeline that nevertheless scores poorly under the source evaluator. Our finding that an oracle-on-demand recovers much of the loss while unaided agents silently commit places this in the selective-prediction and calibrated-abstention tradition~\citep{kamath2020selective,kadavath2022language,zhang2024r}: the capability to defer is present, but its calibration to the right inputs is not. Table~\ref{tab:benchmark_comparison} summarizes this positioning.


\section{Problem Setup}
\label{sec:method}

A data-science agent receives an \emph{observation} $o=(p,d)$ consisting of a prompt $p$ and data package $d$. To act, it must commit to a \emph{framing}: the discretionary choices that turn $o$ into a well-posed supervised-learning problem (target, objective, output form, prediction-time feature availability, permissible external information, etc.). Many such choices exist; we focus on two representative variables, $\theta=(T,M)$, where $T$ specifies the prediction target (which quantity to predict) and $M$ specifies the evaluation objective together with the output form it implies (e.g., calibrated probabilities vs.\ hard labels, ranking vs.\ pointwise scores). We single these out because they are (i)~directly tied to the source-benchmark evaluator, so misframing is scorable without changing the scoring rule; and (ii)~controllable through minimal prompt--data edits that leave the rest of the task intact. Other framing variables are held fixed in this benchmark and left to future work. We write $\theta^\star$ for the intended framing and $\hat\theta$ for the framing the agent commits to. Ambig-DS instantiates $T$ on DSBench (Ambig-DS-Target) and $M$ on MLE-bench (Ambig-DS-Objective); each suite varies one component while the other is fully specified by its source task, so we analyze suites independently and never aggregate or compare effect magnitudes across them.

Let $\mathcal{C}(o)$ be the set of framings admitted by $o$ under the verification protocol $\Pi$ described in \S\ref{sec:verification}: $\Pi$ is a four-item human-and-LLM checklist (plausible alternatives, ambiguity preservation, decision relevance, task preservation). The fully specified source observation $o^\star$ satisfies $\mathcal{C}(o^\star)=\{\theta^\star\}$; a constructed observation is \emph{ambiguous} if $|\mathcal{C}(o)|>1$. Faced with such an observation, the agent must either ask for clarification or commit to some $\hat\theta\in\mathcal{C}(o)$. An \emph{unflagged misframing} occurs when the agent acts without clarification and commits to $\hat\theta\neq\theta^\star$, producing a coherent artifact under an unintended framing. The label $\theta^\star$ is a \emph{protocol convention} inherited from the source task and used purely for scoring: the agent should not recover $\theta^\star$ from an underdetermined observation, but it should recognize the underdetermination and ask before silently committing. A well-calibrated agent satisfies: ask iff $|\mathcal{C}(o)|>1$; commit to $\theta^\star$ when $\mathcal{C}(o)=\{\theta^\star\}$. Ambig-DS evaluates deviations from this rule via paired (Full, Ambig., Ask) conditions.

\paragraph{Two concrete instances.} A \emph{target} instance (\texttt{bike-sharing-demand}, DSBench): the original prompt says ``predict \texttt{count}'' and the data exposes a \texttt{count} column; the ambiguous variant renames it to \texttt{val\_1}, adds a calibrated decoy \texttt{val\_2}, and rewrites the prompt to ``predict value''---both columns are plausible, only \texttt{val\_1} is scored. An \emph{objective} instance (\texttt{aerial-cactus-identification}, MLE-bench): the original prompt evaluates submissions by ROC-AUC and the format is \texttt{id,has\_cactus}; the ambiguous variant keeps the data, target, and submission format identical but replaces the metric sentence with ``a standard evaluation procedure for this task,'' so the agent does not know whether to submit probabilities or hard labels. Full prompt diffs in Appendix~\ref{app:dataset_examples}.

\section{Benchmark Construction}

\begin{figure*}[t]
\centering
\small
\resizebox{0.9\textwidth}{!}{%
\begin{tikzpicture}[
    stage/.style={draw, fill=gray!5, rounded corners=2pt, align=center, inner sep=6pt, text width=0.15\textwidth, minimum height=1.4cm, font=\small},
    detail/.style={draw, rounded corners=2pt, align=left, inner sep=6pt, text width=0.28\textwidth, font=\footnotesize, minimum height=2.2cm},
    arrow/.style={-{Latex[length=2.5mm]}, thick}
]

\node[stage] (s1) {\textbf{1. Source Task}\\ \scriptsize DSBench, MLE-bench};
\node[stage, right=0.6cm of s1] (s2) {\textbf{2. Screen}\\ \scriptsize Executable \& \\ Transformable};
\node[stage, right=0.6cm of s2] (s3) {\textbf{3. Generate}\\ \scriptsize Ambiguous \\ Variants};
\node[stage, right=0.6cm of s3] (s4) {\textbf{4. Verify}\\ \scriptsize LLM + Human \\ Review};
\node[stage, right=0.6cm of s4] (s5) {\textbf{5. Package}\\ \scriptsize Ambiguous + Full \\ w/ Evaluator};

\draw[arrow] (s1) -- (s2); \draw[arrow] (s2) -- (s3); \draw[arrow] (s3) -- (s4); \draw[arrow] (s4) -- (s5);

\node[detail, below=1.0cm of s2, xshift=-1.62cm, fill=blue!2] (d1) {
    \textbf{Target Ambiguity}\\
    \textit{Prompt--data intervention:}\\[0.2em]
    $\bullet$ Remove target cues in prompt\\
    $\bullet$ Add plausible decoy target\\
    $\bullet$ Rename target (\texttt{val\_1}/\texttt{val\_2})\\
};

\node[detail, right=0.4cm of d1, fill=blue!2] (d2) {
    \textbf{Objective Ambiguity}\\
    \textit{Prompt-only intervention:}\\[0.2em]
    $\bullet$ Remove explicit metric refs\\
    $\bullet$ Keep data/submission fixed\\
};

\node[detail, right=0.4cm of d2, fill=green!2] (d3) {
    \textbf{Retention Criteria}\\[0.2em]
    \checkmark Plausible alternatives\\
    \checkmark Ambiguity preserved\\
    \checkmark Decision-relevant ambiguity\\
    \checkmark Original solvability preserved
};

\draw[thick, gray, dashed] (s3.south) -- ++(0,-0.4) -| (d1.north);
\draw[thick, gray, dashed] (s3.south) -- ++(0,-0.4) -| (d2.north);
\draw[thick, gray, dashed] (s4.south) -- (d3.north);

\end{tikzpicture}
}
\caption{\textbf{Ambig-DS construction pipeline.} Tasks are transformed through two pathways: {target ambiguity} (manipulating prompt and data) and {objective ambiguity} (manipulating the prompt only). Final variants are filtered through human and LLM verification to ensure the ambiguity is decision-relevant and the underlying task is preserved.}
\label{fig:construction_pipeline}
\end{figure*}
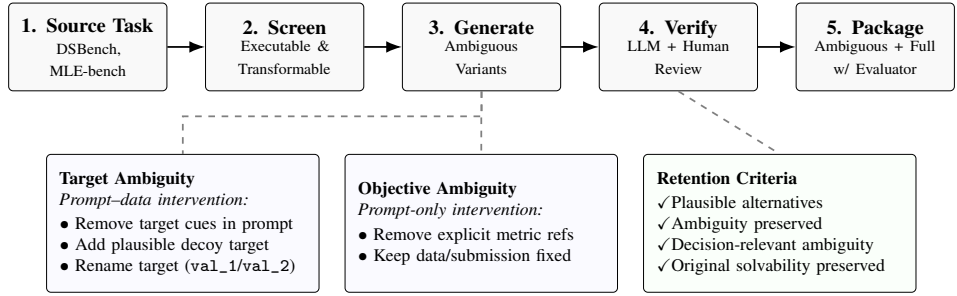

We construct \textbf{Ambig-DS} by converting fully specified Kaggle-style predictive tasks into paired packages: the original task and an ambiguous variant in which the prompt--data observation no longer uniquely determines the intended framing. The ambiguous variant preserves the original evaluator, so performance changes reflect task-framing ambiguity rather than a changed scoring rule. The two suites are built on disjoint source benchmarks chosen for the framing variable they isolate, and we describe their construction independently below; Section~\ref{sec:verification} covers the shared verification and filtering procedure, and Section~\ref{sec:oracle} specifies the clarification oracle that ships with each released task and is invoked under the Ask evaluation condition (\S\ref{sec:experiments}). The five-stage construction pipeline is shown in Figure~\ref{fig:construction_pipeline}; Table~\ref{tab:benchmark_composition} summarizes the retained tasks. Screening retains DSBench tasks with a single clean target column and MLE-bench tasks whose data package fits the harness storage limit (50\,GB). Prompts are in \ref{app:prompts}

\vspace{-5pt}

\subsection{Ambig-DS-Target Construction (DSBench)}

We build the target-ambiguity suite on DSBench because its tabular modeling tasks typically expose a single explicit target column, enabling controlled target interventions while the evaluator is held fixed. We use coordinated prompt--data interventions: we rename the true target column to either \texttt{val\_1} or \texttt{val\_2}, and add a second candidate target with the other name. The true-target ordering is randomized: in the final target set, \texttt{val\_1} is correct in 31 cases and \texttt{val\_2} in 20. The sample submission column headers are replaced with a generic name, e.g., \texttt{id,prediction}, and the test split contains neither candidate target.

The added target is designed to be a plausible alternative, not an obvious distractor. We construct the decoy in two steps. First, we build a vector of scores from a small subset of visible features that are weakly correlated with the true target (in our implementation the bottom $\sim70\%$ of features by $|\rho_{\mathrm{Spearman}}(\cdot, y)|$, clamped to $[4, 40]$). Then we sort training rows by this score vector and assign true-target values in the same sorted order, so the decoy has the same value distribution as the true target but is driven by a different combination of features. To target equal feature-predictability, we add per-task calibrated label noise (label swapping for classification, Gaussian noise injected into the rank-mapped decoy followed by a re-rank-map onto $\mathrm{sort}(y)$ for regression) and binary-search the noise level $\eta$ to drive the 3-fold HistGradientBoosting score of the decoy toward the true target within a target tolerance (AUC for binary classification, accuracy for multiclass, $R^2$ for regression); 29 of 51 retained tasks meet a 0.02 gap, 40 meet 0.05, and 44 meet 0.10. The remaining tasks retain the closest candidate satisfying the marginal and correlation filters. Construction details and validation diagnostics --- exact marginal matches, low target--decoy correlations, small CV-signal gaps, and near-chance inferability under marginal, CV, and LLM schema heuristics --- are in Appendix~\ref{app:decoy}.

We then edit the prompt to remove target-identifying language while preserving the task using Claude Opus 4.7 (worked example: \S\ref{sec:method} and Appendix~\ref{app:example_bike_sharing}). Since both candidate targets appear in training, are absent from the test split, and have plausible marginals and feature signal, the full prompt--data observation does not identify which candidate target is intended.

\subsection{Ambig-DS-Objective Construction (MLE-bench)}

We build the objective-ambiguity suite on MLE-bench because its Kaggle competitions often make the metric central to model selection, thresholding, and submission formatting. We use prompt-only interventions: we keep the data package, target definition, submission format, and evaluator fixed, but remove explicit metric references from the prompt using Claude Opus 4.7. Metric-specific language is replaced with generic performance language (worked example: \S\ref{sec:method} and Appendix~\ref{app:example_aerial_cactus}); the target remains specified while the optimization objective is underspecified.

The metric-ambiguity set includes both higher-is-better and lower-is-better objectives: 27 and 12 tasks, respectively. Metrics include log loss, ROC-AUC variants, word-level Jaccard, accuracy, RMSE/RMSLE, MAE, MAP@K, label-weighted label-ranking average precision, Pearson correlation, Levenshtein rate, and custom geometric distance. This diversity makes metric omission decision-relevant: agents may need to choose hard labels versus probabilities, infer the optimization direction, or account for metric-specific quirks such as top-$K$ ranking, clipping, column-wise averaging, or custom aggregation. See Appendix \ref{app:example_aerial_cactus} for a worked out example.


\subsection{Verification and Filtering}
\label{sec:verification}

The same verification and filtering procedure is applied to both suites. Each candidate passes LLM verification and human review using the same checklist:

\begin{enumerate}
    \item \textbf{Plausible alternatives:} the edited task admits at least two reasonable candidate framings.
    \item \textbf{Ambiguity preservation:} these alternatives remain plausible after considering both the edited prompt and edited data package. This check catches failed edits where a cue removed from one part of the task package (prompt or data) is still revealed elsewhere. 
    \item \textbf{Decision relevance:} resolving the ambiguity changes a task-level choice a competent solver should make, such as the target, output value type, optimization criterion, thresholding, or submission semantics.
    \item \textbf{Task preservation:} the edit removes only information needed to induce the ambiguity; unrelated task information is preserved.
\end{enumerate}

To audit possible construction-model bias, we re-run the verification checklist with non-Claude verifiers and compare their decisions against the final human-reviewed labels; agreement is broadly similar across model families (Appendix~\ref{app:cross_verifier_audit}). This audit tests whether candidate variants that enter the filtering stage satisfy the retention checklist across model families; it does not show that a different generator model would propose the same distribution of candidate variants.

\subsection{Clarification Oracle}
\label{sec:oracle}

Beyond the construction protocol above, each released task ships with a clarification oracle that defines a scoped, truthful channel through which the missing framing variable can be elicited; agents invoke it under the Ask evaluation condition (\S\ref{sec:experiments}). Before modeling, the agent may inspect and explore the data package but cannot fit or evaluate models; it then writes one self-contained question of $\leq 50$ words, or \texttt{NONE}, and stops. Claude Haiku 4.6 answers under a restricted answer-only protocol. For target ambiguity, the oracle answers only the literal question and reveals the target only if explicitly asked; it may not reveal test statistics, cross-validation diagnostics, construction internals, code, or modeling advice. For objective ambiguity, the oracle is restricted to metric, scoring, and submission-format information (metric name, direction, required submission value type, metric quirks). Out-of-scope requests receive a fixed refusal. The oracle is not a simulation of stakeholder interaction; it is a causal probe that asks whether the agent can elicit the missing framing information when a truthful, scoped channel is available, isolating framing recognition from realistic-user variability. Two idealizations cut in opposite directions: one-shot understates what richer multi-turn clarification could recover, while truthful, scoped, always-available answers overstate what noisy real stakeholders would yield. Ask is therefore a causal diagnostic for whether the missing framing information suffices when delivered perfectly, not a deployment estimate. Oracle-family effects (Claude Haiku 4.6) cannot be ruled out; the answer-only protocol restricted to manifest fields mitigates this.

\section{Experiments}
\label{sec:experiments}

\subsection{Experimental Setup}

\paragraph{Agent framework and models.}
We evaluate code-based agents in the OpenCode framework \citep{anomalyco2026opencode}, which supports iterative code generation, execution, debugging, and artifact submission over a prompt--data package. We evaluate efficient models 
(Gemini~3~Flash, GPT-5.4~Nano, Claude~Haiku~4.5) to frontier-class models 
(Gemini~3.1~Pro, GPT-5.4), to characterize how ambiguity sensitivity scales 
with frontier capability. All models are evaluated with the same tool access, execution environment, and per-suite wall-clock budget (2h on Ambig-DS-Target, 24h on Ambig-DS-Objective), the oracle round-trip is not counted towards the budget. Prompts are in \ref{app:prompts}.

\paragraph{Evaluation conditions.}
Each task is evaluated under three conditions. \textbf{Full} is the original fully specified task, with the target and evaluation objective explicit. \textbf{Ambig.} is the ambiguous variant with no clarification oracle available; the agent is not invited to ask, but is also not prevented from flagging the underdetermination or abstaining. An alternative-framing submission therefore reflects a silent unilateral commitment under ambiguity. \textbf{Ask} is the ambiguous variant with the clarification oracle described in Section~\ref{sec:oracle}: before modeling, the agent may ask exactly one self-contained question under 50 words, or output \texttt{NONE}, and then stops until the oracle answer is returned.

\paragraph{Task performance.}
Task success is scored with the original source benchmark evaluator (DSBench score for target ambiguity; MLE-bench competition score for objective ambiguity), normalized to $[0,1]$. Let $S_{\text{full}}$, $S_{\text{ambig}}$, and $S_{\text{ask}}$ denote the per-task normalized scores; we define $\Delta_{\text{ambig}} = S_{\text{ambig}} - S_{\text{full}}$ and $\Delta_{\text{ask}} = S_{\text{ask}} - S_{\text{ambig}}$, macro-averaged across tasks. Negative $\Delta_{\text{ambig}}$ indicates degradation; positive $\Delta_{\text{ask}}$ indicates recovery. Invalid and Timeout runs score $0$ and remain in the denominator (51 Target / 61 Objective tasks per cell). We report one-sided paired Wilcoxon signed-rank tests (Full $>$ Ambig.; Ask $>$ Ambig.) and paired bootstrap 95\% CIs (Appendices~\ref{app:significance},~\ref{app:bootstrap}). All comparisons are within-task and matched across conditions.

\paragraph{Ask behavior.}
We report two ask rates in the ask-enabled setting: \textbf{Ask on Ambig.} (fraction of ambiguous tasks where the agent asks; asking is the desired behavior) and \textbf{Ask on Full} (fraction of fully specified tasks where the agent asks despite the framing being specified). An \emph{ask} is any non-\texttt{NONE} question written to \texttt{\_question.txt}; out-of-scope questions count as asks (and receive the fixed refusal), and are sub-classified in Tables~\ref{tab:over-asking}--\ref{tab:under-asking}. Per-cell denominators are reported in Table~\ref{tab:ask_policy_sensitivity_axis}.

\paragraph{Framing diagnostics.}
We separate task-framing failures from generic execution failures by first checking whether the run produced a parseable submission accepted by the evaluator. For target ambiguity, valid submissions are labeled by whether they optimize the intended or alternative target. For objective ambiguity, five LLM-judge calls with majority vote classify the code, logs, response, and submission as intended-objective, alternative-objective, degenerate, or invalid/no-score; a human verifies the classification. Only valid alternative-framing runs are counted as unflagged misframings. Inter-judge agreement, human override rates, and a cross-family judge/oracle robustness check on Claude-evaluated runs are reported in Appendix~\ref{app:judge_reliability}; headline conclusions are unchanged under either judge or oracle family.

\begin{table*}[t]
\centering
\small
\setlength{\tabcolsep}{4pt}
\resizebox{0.9\textwidth}{!}{%
\begin{tabular}{ll|ccc|cc|cc}
\toprule
\textbf{Axis}
& \textbf{Model}
& \multicolumn{3}{c|}{\textbf{Performance} $\uparrow$}
& \multicolumn{2}{c|}{\textbf{Change}}
& \multicolumn{2}{c}{\textbf{Wilcoxon $p$-value}} \\
\cmidrule(lr){3-5}
\cmidrule(lr){6-7}
\cmidrule(lr){8-9}
& 
& Full
& Ambig.
& Ask
& $\Delta_{\text{ambig}}$ $\downarrow$
& $\Delta_{\text{ask}}$ $\uparrow$
& Full $>$ Ambig.
& Ask $>$ Ambig. \\
\midrule
Objective & Gemini 3 Flash    & 0.31 & 0.22 & 0.31 & $-$0.09 & +0.09 & $<$0.001$^{***}$ & 0.002$^{**}$  \\
Objective & GPT-5.4 Nano      & 0.11 & 0.05 & 0.10 & $-$0.06 & +0.05 & 0.004$^{**}$     & 0.004$^{**}$  \\
Objective & Claude Haiku 4.5  & 0.27 & 0.23 & 0.26 & $-$0.04 & +0.03 & 0.012$^{*}$            & 0.010$^{*}$        \\
Objective & Gemini 3.1 Pro & 0.27 & 0.23 & 0.27 & $-$0.04 & +0.04 & 0.011$^{*}$  & 0.012$^{*}$  \\
Objective & GPT-5.4           & 0.26 & 0.22 & 0.26 & $-$0.04 & +0.04 & 0.004$^{**}$     & 0.021$^{*}$   \\
\midrule
Target & Gemini 3 Flash       & 0.63 & 0.44 & 0.61 & $-$0.19 & +0.17 & 0.018$^{*}$      & 0.021$^{*}$   \\
Target & GPT-5.4 Nano         & 0.49 & 0.39 & 0.45 & $-$0.10 & +0.06 & 0.002$^{**}$     & 0.020$^{*}$   \\
Target & Claude Haiku 4.5     & 0.56 & 0.32 & 0.52 & $-$0.24 & +0.20 & $<$0.001$^{***}$ & 0.002$^{**}$  \\
Target & Gemini 3.1 Pro       & 0.64 & 0.49 & 0.60 & $-$0.15 & +0.11 & $<$0.001$^{***}$ & $<$0.001$^{***}$ \\
Target & GPT-5.4              & 0.61 & 0.32 & 0.52 & $-$0.29 & +0.20 & $<$0.001$^{***}$ & $<$0.001$^{***}$ \\
\bottomrule
\end{tabular}
}
\caption{\textbf{Main results.} Target ambiguity on DSBench, objective ambiguity on MLE-bench. Scores are normalized to $[0,1]$. We report per-condition performance, paired changes ($\Delta_{\text{ambig}}$, $\Delta_{\text{ask}}$), and one-sided Wilcoxon tests (Full $>$ Ambig.; Ask $>$ Ambig.). Bootstrap 95\% CIs are in Table~\ref{tab:bootstrap_cis}.}
\vspace{-0.5em}
\label{tab:axis_results}
\end{table*}

\subsection{Results: Ambig-DS-Target (DSBench)}

\paragraph{Frontier agents silently commit to a wrong target.}
Even frontier backbones (Gemini~3.1~Pro, GPT-5.4) submit a plausible alternative target on $39$--$63\%$ of ambiguous DSBench runs without flagging the underdetermination (Table~\ref{tab:target_diagnostic}); the smaller backbones misframe at $43$--$59\%$. These valid alternative-target submissions are \emph{unflagged misframings} (Section~\ref{sec:method}). Appendix~\ref{app:case_studies} (\texttt{bike-sharing-demand}) shows an agent inventing a plausible but unsupported target composition rather than asking for clarification, and a trace/answer recognition audit (Appendix~\ref{app:recognition_audit}) shows that user-overridable flagging in the final answer never exceeds $4\%$ even when traces acknowledge the ambiguity, so the silent-commitment characterization holds at the answer-facing level. Conditional on category, intended runs score near Full and alternative-target runs collapse (Appendix~\ref{app:score_by_category}).

\paragraph{This translates to score loss.}
On Target rows of Table~\ref{tab:axis_results}, $\Delta_{\text{ambig}}$ ranges from $-0.10$ to $-0.29$; all five bootstrap CIs exclude zero (Table~\ref{tab:bootstrap_cis}; tightest GPT-5.4~Nano, $[-0.18,-0.02]$); all paired Wilcoxon tests for Full $>$ Ambig.\ are significant.

\begin{table*}[t]
\centering
\small
\setlength{\tabcolsep}{5pt}
\resizebox{0.9\textwidth}{!}{%
\begin{tabular}{l|cc|cc|cc|cc}
\toprule
\textbf{Model}
& \multicolumn{2}{c|}{\textbf{Intended Target} $\uparrow$}
& \multicolumn{2}{c|}{\textbf{Alternative Target} $\downarrow$}
& \multicolumn{2}{c|}{\textbf{Invalid} $\downarrow$}
& \multicolumn{2}{c}{\textbf{Timeout} $\downarrow$} \\
\cmidrule(lr){2-3}
\cmidrule(lr){4-5}
\cmidrule(lr){6-7}
\cmidrule(lr){8-9}
& Ambig. & Ask
& Ambig. & Ask
& Ambig. & Ask
& Ambig. & Ask \\
\midrule
Gemini 3 Flash   & 43\% & 96\% & 43\% & 2\%  & 6\% & 0\% & 8\%  & 2\%  \\
GPT-5.4 Nano     & 47\% & 78\% & 47\% & 18\% & 4\%  & 2\% & 2\%  & 2\%  \\
Claude Haiku 4.5 & 26\% & 88\% & 59\% & 4\%  & 8\% & 4\% & 8\%  & 4\%  \\
Gemini 3.1 Pro   & 55\% & 96\% & 39\% & 4\%  & 4\%  & 0\% & 2\%  & 0\%  \\
GPT-5.4          & 27\% & 71\% & 63\% & 24\% & 6\% & 2\% & 4\%  & 4\%  \\
\bottomrule
\end{tabular}
}
\caption{\textbf{Target-framing diagnostics.} Each run is classified as optimizing the intended target, an alternative plausible target, or yielding no valid score. Under \textbf{Ambig.} the agent is not invited to ask, so alternative-target submissions are unilateral commitments to an unintended target without acknowledging the underdetermination---i.e., \emph{unflagged misframings}. \emph{Invalid} (missing/unparseable/rejected) and \emph{Timeout} (budget exhausted) are reported separately and not counted as misframings.}
\vspace{-0.5em}
\label{tab:target_diagnostic}
\end{table*}

\subsection{Results: Ambig-DS-Objective (MLE-bench)}

\paragraph{Ambiguity degrades performance.}
On Ambig-DS-Objective, $\Delta_{\text{ambig}}$ ranges from $-0.04$ to $-0.09$ (15--35\% of Full); all five bootstrap CIs exclude zero (Table~\ref{tab:bootstrap_cis}; tightest GPT-5.4, $[-0.06,-0.02]$); all paired Wilcoxon tests for Full $>$ Ambig.\ are significant.

\paragraph{Silent failures take two forms: misframing and abdication.}
Table~\ref{tab:objective_diagnostic} shows that the dominant Ambiguous-condition failures split between two modes that share a silent-failure signature. \emph{Unflagged misframing} (\textit{Alternative Obj.}) is a substantive model optimized for a plausible but unintended objective. \emph{Unflagged abdication} (\textit{Degenerate}) is a non-committal submission such as a copied baseline or constant prediction. Both produce valid, scoreable artifacts; neither acknowledges the underdetermination or asks for clarification; both are quietly suboptimal under the intended objective. Combined silent failure runs $16$--$62\%$ under Ambig.\ across all backbones (frontier: $46\%$ Gemini~3.1~Pro, $19\%$ GPT-5.4), splitting into $5$--$18\%$ misframing and $8$--$49\%$ abdication. Under Ask, \emph{Alternative Obj.}\ drops to $\leq 4\%$ everywhere, but \emph{Degenerate} persists for weaker backbones --- Ask kills misframing, not abdication. Ask also reduces Invalid submissions, plausibly because some format mismatches resolve once the oracle reveals the required submission value type. Appendix~\ref{app:case_studies} (\texttt{random-acts-of-pizza}) shows an agent treating an AUC probability-ranking task as hard binary classification when the metric is omitted. Same conditional pattern holds, with degenerate runs near zero (Appendix~\ref{app:score_by_category}).

\begin{table*}[t]
\centering
\small
\setlength{\tabcolsep}{5pt}
\resizebox{0.9\textwidth}{!}{%
\begin{tabular}{l|cc|cc|cc|cc|cc}
\toprule
\textbf{Model}
& \multicolumn{2}{c|}{\textbf{Intended Obj.} $\uparrow$}
& \multicolumn{2}{c|}{\textbf{Alternative Obj.} $\downarrow$}
& \multicolumn{2}{c|}{\textbf{Degenerate} $\downarrow$}
& \multicolumn{2}{c|}{\textbf{Invalid} $\downarrow$}
& \multicolumn{2}{c}{\textbf{Timeout} $\downarrow$} \\
\cmidrule(lr){2-3}
\cmidrule(lr){4-5}
\cmidrule(lr){6-7}
\cmidrule(lr){8-9}
\cmidrule(lr){10-11}
& Ambig. & Ask
& Ambig. & Ask
& Ambig. & Ask
& Ambig. & Ask
& Ambig. & Ask \\
\midrule
Gemini 3 Flash   & 75\% & 84\% & 5\%  & 0\% & 11\% & 11\% & 7\% & 2\% & 2\%  & 3\%  \\
GPT-5.4 Nano     & 38\% & 54\% & 13\%  & 0\% & 49\% & 46\% & 0\% & 0\% & 0\%  & 0\%  \\
Claude Haiku 4.5 & 54\% & 95\% & 18\%  & 0\% & 21\% & 5\%  & 5\% & 0\% & 2\%  & 0\%  \\
Gemini 3.1 Pro   & 51\% & 59\% & 13\%  & 3\% & 33\% & 33\% & 2\% & 3\% & 2\%  & 2\%  \\
GPT-5.4          & 72\% & 79\% & 11\% & 0\% & 8\%  & 11\% & 5\% & 7\% & 4\%  & 4\%  \\
\bottomrule
\end{tabular}
}
\caption{\textbf{Objective-framing diagnostics.} Majority vote over five LLM-judge calls on each run's trajectory and submission. \emph{Intended Obj.}: substantive model aligned with the source metric. \emph{Alternative Obj.}: substantive model optimized for a plausible but unintended objective. \emph{Degenerate}: no real modeling attempt (e.g., copied baseline, constants). \emph{Invalid}: missing/unparseable; \emph{Timeout}: budget exhausted. Invalid and Timeout are split to surface wall-clock confounds.}
\vspace{-0.5em}
\label{tab:objective_diagnostic}
\end{table*}

\subsection{Cross-Suite Observations}
The degradation-under-Ambig.\ and recovery-under-Ask pattern holds in both suites and is not specific to Claude-family evaluated agents: restricting to the four non-Claude backbones in Table~\ref{tab:axis_results} (Gemini~3~Flash, Gemini~3.1~Pro, GPT-5.4~Nano, GPT-5.4) preserves the qualitative pattern within each suite --- every non-Claude row shows $\Delta_{\text{ambig}} < 0$ and $\Delta_{\text{ask}} > 0$ with all paired Wilcoxon tests significant at $p<0.05$. To rule out same-family inflation in the Ask condition, we re-run Claude Haiku 4.5 Ask with a GPT-5.4 oracle: $\Delta_{\text{ask}}$ shifts $+0.20\to+0.18$ on Target, unchanged on Objective. Swapping the Objective judge from Claude to GPT shifts per-cell misframing by $\leq 4$pp ($\kappa{=}0.74$); suggesti g recovery is not an oracle- or judge-family artifact (Appendices~\ref{app:cross_verifier_audit},~\ref{app:judge_reliability}).

\begin{figure*}[t]
\centering
\includegraphics[width=0.7\textwidth]{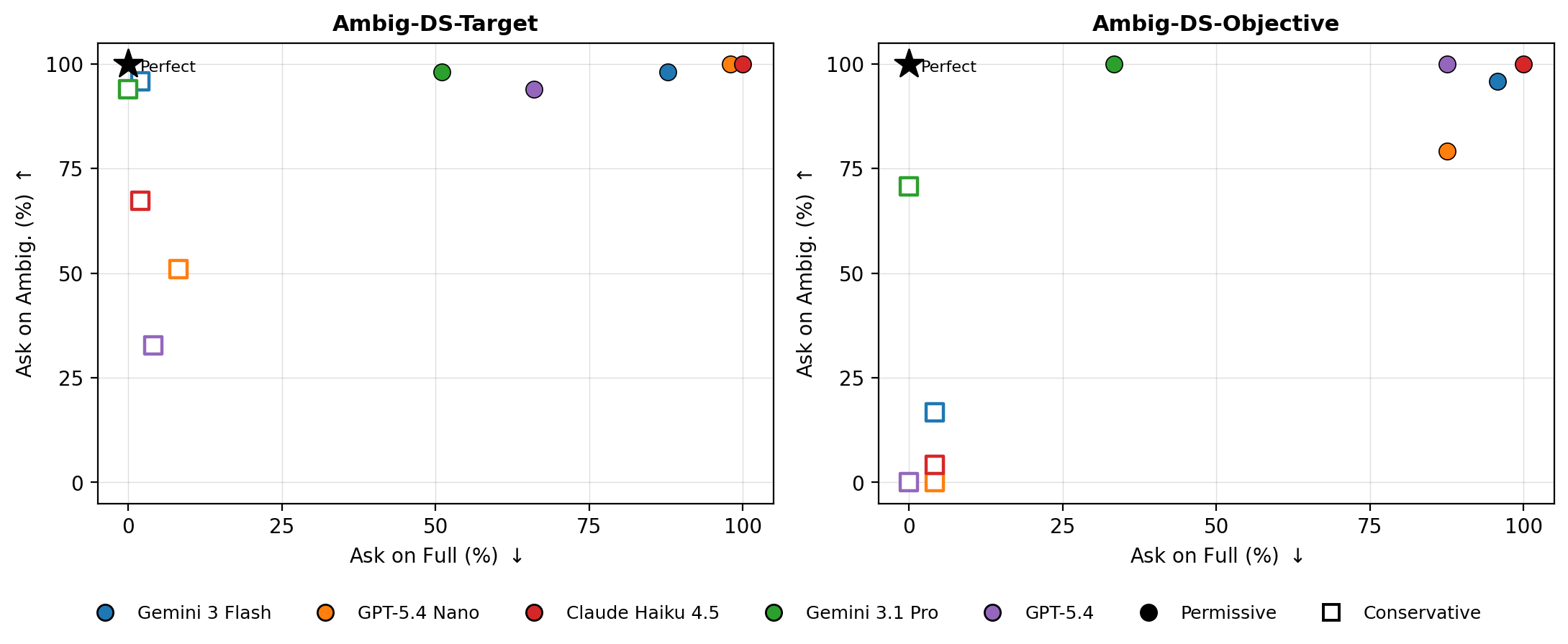}
\caption{\textbf{Ask-policy sensitivity.} Each point is one (model, policy) on one suite. Vertical axis: clarification recall on ambiguous tasks ($\uparrow$). Horizontal axis: unnecessary clarification on fully specified tasks ($\downarrow$). Perfect calibration is the top-left $\star$. \emph{Permissive} (filled circle) yields high recall but high false positives; \emph{conservative} (open square) reduces over-asking but suppresses legitimate asking, especially on Ambig-DS-Objective. Numerical values are in Table~\ref{tab:ask_policy_sensitivity_axis}.}
\label{fig:ask_policy_sensitivity}
\end{figure*}

\paragraph{Agents can use clarification when obtained, but cannot reliably detect when to ask.}
Appendix~\ref{app:ask_act_failure_analysis} reports the full ask--act 
calibration analysis. 
Calibration varies 
substantially across models and policies within each suite (Figure~\ref{fig:ask_policy_sensitivity}; numerical values in Table~\ref{tab:ask_policy_sensitivity_axis}): on Ambig-DS-Target, some model--policy combinations approach good calibration (e.g., Gemini~3.1~Pro under the conservative policy asks on $94\%$ of ambiguous tasks and $0\%$ of fully specified ones), while on Ambig-DS-Objective most agents collapse under the conservative policy (Claude~Haiku~4.5 asks on only $5\%$ of ambiguous metric tasks; GPT-5.4~Nano and GPT-5.4 ask on $0\%$). Within each suite, the permissive policy yields high recall on 
ambiguous tasks at the cost of high false positives on fully specified ones, 
while the conservative policy reduces over-asking but suppresses legitimate 
asking in turn.

The diagnostics reveal what goes wrong at each end. Over-asking is rarely a 
legitimate response to missing task information: on fully specified tasks under 
the permissive policy, unnecessary questions are dominated by style preferences, 
restatements of information already present in the prompt, and out-of-scope
requests for modeling guidance --- e.g., $88\%$ of Claude~Haiku~4.5's asks on Target are style-preference questions and $85\%$ of Gemini~3~Flash's asks on Objective are out-of-scope requests (Table~\ref{tab:over-asking}). Under-asking manifests not as asking the wrong question but as silently committing to a default framing without acknowledging the ambiguity, particularly in Ambig-DS-Objective where agents proceed without the metric rather than flagging its absence --- under the conservative policy, GPT-5.4 and GPT-5.4~Nano silently default on $100\%$ of ambiguous metric runs and Claude~Haiku~4.5 on $97\%$, while $0$--$8\%$ of asks ever target the wrong topic (Table~\ref{tab:under-asking}). Permissive and conservative bracket the practical operating range; observing that \emph{both} yield miscalibration on at least one suite for every model--policy pair we tested\footnote{The calibration claim is over the 5 models $\times$ 2 policies $\times$ 2 suites = 20 cells we evaluate; the released task packages, evaluators, oracle protocol, and judge prompts are public, so extending this audit to additional (in particular open-weights) backbones requires only re-running the harness (App.~\ref{app:extended_limitations}).} is consistent with the qualitative claim that the failure is the absence of reliable internal detection, which no prompt policy in our study substitutes for.

To assess whether asking on ambiguous tasks reflects genuine framing recognition, we examine recovery under the clarification oracle. When the missing framing variable is elicited and incorporated, much of the ambiguity-induced degradation is recoverable within each suite, with $\Delta_{\text{ask}}$ reaching $+0.20$ on Ambig-DS-Target and $+0.09$ on Ambig-DS-Objective (Table~\ref{tab:axis_results}); Ask essentially closes the Ambig.--Full gap on Objective but leaves a residual gap on Target. Agents have the capacity to integrate framing information when provided perfectly; whether they can extract it from realistic stakeholder interaction is untested by this protocol. Ask does not fully recover Full for most model--suite pairs, so residual execution failures persist even after framing is resolved.

\vspace{-5pt}

\section{Discussion and Limitations}
\vspace{-5pt}
Ambig-DS isolates a failure mode that standard data-science benchmarks largely hide. In fully specified settings, a valid submission is often taken as evidence that the agent understood the task. Under task-framing ambiguity, this inference breaks down: an agent can produce a coherent artifact while committing to a plausible but unintended target or objective. This separates pipeline-execution competence from task-framing competence.

The two suites yield distinct silent-failure signatures. Ambig-DS-Target produces discrete wrong-target misframings; Ambig-DS-Objective splits between substantive misframings (commitment to a plausible alternative objective) and abdications (retreat to a degenerate baseline or constant submission). Both are unflagged commitments under ambiguity rather than execution failures.

Within each suite, the clarification oracle recovers much of the lost performance, indicating that missing framing information --- not pipeline execution --- drives the degradation. The non-trivial finding is not that recovery occurs but that agents cannot reliably detect when to invoke clarification. 

Ambig-DS yields actionable signals for three audiences. \textit{Users} should specify framing variables (target, metric, output type, submission semantics) explicitly: omitting them produces reliable and often large performance drops even from capable agents. \textit{Builders} should treat ask--act calibration as a concrete post-training or prompt-engineering target, since prompt wording shifts the operating point but does not yield robust sensitivity to decision-relevant ambiguity. \textit{Benchmark designers} should pair fully specified tasks with diagnostic ambiguous variants: pipeline-execution scores on fully specified tasks overestimate competence when framing is not guaranteed.

\vspace{-5pt}

\paragraph{Limitations.}

Ambig-DS is a controlled diagnostic suite, not a census. We construct minimal interventions on existing benchmark tasks while preserving the original evaluator, and study only two framing variables ($T$, $M$); other framing variables (information boundaries, data leakage, temporal assumptions, prediction-time feature availability, permissible external information) and richer naturalistic ambiguity are kept for future work. Axis is confounded with source benchmark by design: target ambiguity is instantiated on DSBench (clean tabular targets), objective ambiguity on MLE-bench (metric-sensitive competitions), so effect-size differences across suites are not a causal comparison and we interpret each suite independently. Construction and oracle are model-assisted (Claude Opus generation/verification, Claude Haiku oracle); we mitigate via cross-family verifier audit (App.~\ref{app:cross_verifier_audit}), cross-family judge/oracle robustness (App.~\ref{app:judge_reliability}), human review, and the non-Claude agent subset, but residual generation- and oracle-family effects cannot be fully ruled out. Finally, $\theta^\star$ is a protocol scoring convention rather than the unique correct reading; the Ask oracle is intentionally idealized and should be read as an upper bound on recoverable performance; the dataset is modest in size; and because the Ambig.\ condition has no clarification channel, $\Delta_{\text{ambig}}$ does not fully separate non-recognition from commitment-under-recognized-ambiguity, a confound bounded at the trace level by the recognition audit (Appendix~\ref{app:recognition_audit}). Together these mean Ambig-DS estimates should be read as diagnostic evidence of failure modes under verified ambiguity rather than as prevalence over the space of all data-science tasks; full discussion in Appendix~\ref{app:extended_limitations}.

\bibliographystyle{plainnat}
\bibliography{bib}

\appendix

\section{Extended Limitations}
\label{app:extended_limitations}

This appendix expands the main-body Limitations into per-item paragraphs. The main paper retains the four-paragraph compressed version; reviewers seeking a granular treatment of any individual concern should consult the matching paragraph below.

\paragraph{Naturalistic ambiguity not captured.}
Ambig-DS uses minimal, controlled prompt--data interventions on existing benchmark tasks so that performance changes can be attributed causally to task-framing ambiguity. This isolation is deliberate but does not capture the full messiness of naturally occurring data-science ambiguity, where prompts, legacy datasets, stakeholder context, organizational constraints, and multiple ambiguity types interact. Ambig-DS measures whether agents recognize and act on a single, controlled, decision-relevant ambiguity in an otherwise clean task; it does not measure how agents would behave in fully naturalistic underspecified workflows.

\paragraph{Scope of framing variables ($T$, $M$).}
Ambig-DS deliberately studies only two framing variables: the prediction target $T$, and the evaluation objective $M$. We chose these because they are tied directly to the source-benchmark evaluator (so misframing is scorable without changing the scoring rule) and controllable through minimal prompt--data edits. Other framing variables are also important but are held fixed in this benchmark: information-boundary ambiguity, data leakage, temporal assumptions, prediction-time feature availability, and permissible use of external information. Extending the diagnostic protocol to these axes is future work.

\paragraph{Intended framing as a protocol scoring convention.}
The label $\theta^\star$ inherited from the source task is a protocol convention used for scoring (\S\ref{sec:method}), not the unique correct reading of the ambiguous observation. The diagnostic claim is not that the agent should recover $\theta^\star$ from an underdetermined observation, but that it should recognize the underdetermination and ask before silently committing. The reported numbers therefore measure unflagged commitment under ambiguity against a fixed scoring convention; they do not measure whether the agent's chosen alternative framing is itself reasonable in absolute terms.

\paragraph{Idealized clarification oracle.}
Beyond the construction protocol above, each released task ships with a clarification oracle that defines a scoped, truthful channel through which the missing framing variable can be elicited; agents invoke it under the Ask evaluation condition (\S\ref{sec:experiments}). Before modeling, the agent may inspect and explore the data package but cannot fit or evaluate models; it then writes one self-contained question of $\leq 50$ words, or \texttt{NONE}, and stops. Claude Haiku 4.6 answers under a restricted answer-only protocol. For target ambiguity, the oracle answers only the literal question and reveals the target only if explicitly asked; it may not reveal test statistics, cross-validation diagnostics, construction internals, code, or modeling advice. For objective ambiguity, the oracle is restricted to metric, scoring, and submission-format information (metric name, direction, required submission value type, metric quirks). Out-of-scope requests receive a fixed refusal. The oracle is not a simulation of stakeholder interaction; it is a causal probe that asks whether the agent can elicit the missing framing information when a truthful, scoped channel is available, isolating framing recognition from realistic-user variability. Two idealizations apply in opposite directions: the one-shot protocol understates what richer multi-turn clarification could recover, while truthful, always-available answers overstate what noisy real stakeholders would yield. We therefore read $\Delta_{\text{ask}}$ as a causal diagnostic for whether missing framing information suffices to recover performance when delivered perfectly, not as a deployment estimate. Oracle-family effects (Claude Haiku 4.6) cannot be ruled out; we mitigate by constraining answers to manifest fields under the answer-only protocol.

\paragraph{Axis--source-benchmark confound.}
Axis is confounded with source benchmark by design: target ambiguity is instantiated on DSBench, while objective ambiguity is instantiated on MLE-bench. This pairing is principled---DSBench exposes clean tabular targets that allow controlled target interventions, while MLE-bench contains Kaggle competitions where the metric is central to model selection, thresholding, and submission formatting---but it means effect-size differences across the two suites should not be interpreted as a causal comparison between target and objective ambiguity. Throughout the paper we report and interpret each suite independently and emphasize the qualitative failure signatures within each axis. Full cross-benchmark ablations (e.g., target ambiguity on MLE-bench-style tasks where feasible, objective ambiguity on DSBench tasks with decision-relevant metric variation) are left to future work.

\paragraph{Construction- and oracle-family bias.}
Benchmark construction is model-assisted: Claude Opus 4.7 proposes prompt edits and supports verification, and Claude Haiku 4.6 verbalizes oracle answers under the Ask condition while Claude Haiku 4.5 also appears as an evaluated backbone. This could introduce model-family bias in two places. First, the cross-verifier audit (Appendix~\ref{app:cross_verifier_audit}) tests whether retained candidates satisfy the checklist across non-Claude verifiers, but it does not prove that non-Claude generators would have proposed the same candidate distribution. Second, although the oracle is constrained to short, manifest-grounded answers under the answer-only protocol, oracle phrasing could still interact with agent instruction-following behavior. We mitigate by inheriting $\theta^\star$ from source benchmarks, scoring with the original evaluators, applying human review, auditing with GPT- and Gemini-family verifiers, swapping the oracle to GPT-5.4 for the same-family-attack-sharpest agent (Appendix~\ref{app:judge_reliability}), and confirming that the degradation/recovery pattern holds for non-Claude agents. Generation- and oracle-family effects nonetheless cannot be fully ruled out.

\paragraph{Dataset size and prevalence interpretation.}
The dataset is modest in size (51 paired Target tasks, 61 paired Objective tasks). Combined with the controlled-interventions design, Ambig-DS estimates should be read as diagnostic evidence of failure modes under verified ambiguity, not as precise prevalence estimates over the population of all data-science tasks. Confidence intervals (Appendix~\ref{app:bootstrap}) and within-suite paired tests (Appendix~\ref{app:significance}) quantify the uncertainty of the diagnostic effects we do report.

\paragraph{Open-weights replication.}
Our model selection prioritizes capability range (efficient and frontier-class API models from three families) over weight access. We do not include open-weights backbones in the headline tables. Because the released task packages, evaluators, oracle protocol, and judge prompts are public, replicating Ambig-DS on open-weights agents requires only re-running the evaluation harness; we view this as a natural next step rather than a limitation of the benchmark itself.

\paragraph{Recognition--commitment confound under Ambig.}
The Ambiguous condition has no clarification channel by construction (\S\ref{sec:experiments}), so $\Delta_{\text{ambig}}$ cannot fully disentangle two failure modes: (i) the agent fails to recognize the underdetermination, and (ii) the agent recognizes it but commits silently anyway. The recognition audit (Appendix~\ref{app:recognition_audit}) bounds this confound at the trace level: user-overridable flagging in the final answer never exceeds $4\%$ on Ambig-DS-Target even when traces acknowledge the ambiguity, so the silent-commitment characterization holds at the answer-facing level on Target; on Objective the confound is empirically negligible.

\begin{figure*}[t]
\centering
\resizebox{0.9\textwidth}{!}{%
\begin{tikzpicture}[
    box/.style={
        draw,
        rounded corners=2pt,
        align=left,
        inner sep=5pt,
        text width=0.25\textwidth,
        minimum height=3.8cm,
        font=\footnotesize
    },
    failbox/.style={
        draw=red!70!black,
        fill=red!2,
        rounded corners=2pt,
        align=left,
        inner sep=5pt,
        text width=0.26\textwidth,
        minimum height=3.8cm,
        font=\footnotesize
    },
    arrow/.style={-{Latex[length=2mm]}, thick}
]

\node[box] (full) {
    \textbf{Fully Specified Task}\\ \rule{\linewidth}{0.3pt} \\
    \textbf{Prompt:} predict \texttt{count}\\
    \textbf{Data:} \{\dots, \texttt{count}\}\\
    \textbf{Submission Format:} \texttt{ID, count}\\[1.4em]
    \textit{Framing determined from\\local cues.}
};

\node[box, right=1.3cm of full] (ambig) {
    \textbf{Ambiguous Observation}\\ \rule{\linewidth}{0.4pt} \\
    \textbf{Prompt:} predict \texttt{value}\\
    \textbf{Data:} \{\dots, \texttt{val\_1, val\_2}\}\\
    \textbf{Submission Format:} \texttt{ID, prediction}\\[0.6em]
    \textit{\texttt{val\_1} and \texttt{val\_2} are both plausible targets}
};

\node[failbox, right=1.3cm of ambig] (fail) {
    \textbf{Act Without Clarification}\\ \rule{\linewidth}{0.4pt} \\
    \textbf{Agent Action:}\\
    $\bullet$ Commits to target = \texttt{val\_2}\\
    $\bullet$ Produces valid pipeline\\[0.5em]
    \textbf{Ground Truth:}\\
    $\bullet$ Evaluator expects \texttt{val\_1}\\[1.2em]
    \centering \textbf{\color{red!70!black} \small $\rightarrow$ \textit{Unflagged Misframing}}
};

\draw[arrow] (full) -- node[midway, above, font=\tiny, align=center, yshift=2pt] {remove \\ framing cues} (ambig);
\draw[arrow] (ambig) -- node[midway, above, font=\tiny, align=center, yshift=2pt] {act without\\asking} (fail);

\end{tikzpicture}
}
\caption{\textbf{Conceptual overview: from specified task to unflagged misframing.} Ambig-DS converts fully specified tasks (left) into ambiguous observations that remain executable (middle). An agent acting without clarification may produce a technically valid pipeline under an unintended framing, resulting in an unflagged misframing (right).}
\label{fig:framing_ambiguity}
\end{figure*}
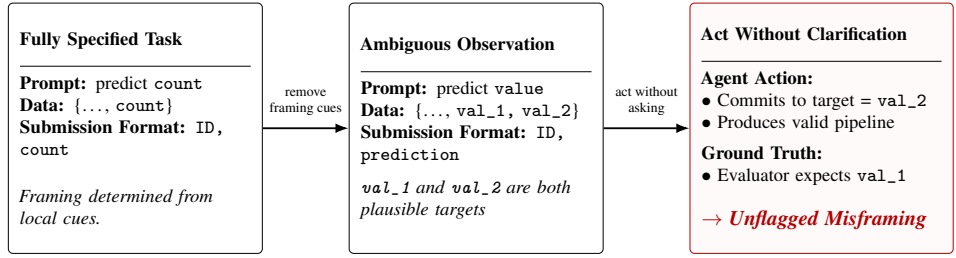

\section{LLM Usage Statement}

We used large language models to assist with writing-level edits, including polishing prose, condensing text, improving clarity, formatting tables, and identifying relevant citations. We also used coding agents for implementation support, debugging, and experiment management. In addition, LLMs were used as part of the dataset construction pipeline for generating candidate ambiguous variants, verifying candidate tasks under a fixed checklist, answering clarification questions under a restricted oracle protocol, and judging objective-framing diagnostics as described in the paper. All research questions, benchmark design decisions, experimental methodology, analyses, and conclusions were developed and validated by the authors. The authors reviewed and approved all generated text, code, prompts, annotations, and experimental outputs used in the paper.

\section{Additional Experimental Details}

\subsection{Benchmark Comparison and Composition}
Figure~\ref{fig:framing_ambiguity} provides a conceptual overview of how Ambig-DS converts a fully specified task into an ambiguous observation that can elicit an unflagged misframing. Table~\ref{tab:benchmark_comparison} shows the comparison of Ambig-DS to other benchmarks and Table~\ref{tab:benchmark_composition} shows the benchmark composition.

\begin{table*}[t]
\centering
\small
\resizebox{1\textwidth}{!}{%
\begin{tabular}{lccccc}
\toprule
Benchmark 
& Executable Task
& Predictive DS/ML 
& Controlled Ambig. 
& Ask--Act Eval. 
& \# Tasks for ML/DS
\\
\midrule
DSBench \cite{jing2024dsbench} & \checkmark & \checkmark & -- & -- & 74 \\
MLE-bench \cite{chan2024mle} & \checkmark & \checkmark & -- & -- & 75 \\
DataSciBench \cite{zhang2025datascibench} & \checkmark & \checkmark & -- & --& 222\\
CoTA \cite{li2025large} & partial & -- & partial & partial & -- \\
Gomm et al. \cite{gomm2025we} & -- & -- & \checkmark & -- & -- \\
AmbiQT \cite{bhaskar2023benchmarking} & \checkmark & -- & \checkmark & -- & --\\
ClarEval \cite{li2026clareval} & \checkmark & -- & \checkmark & \checkmark & --\\
Ambig-SWE \cite{vijayvargiya2026ambig} & \checkmark & -- & \checkmark & \checkmark & 500 \\
\textbf{Ambig-DS} (ours) & \checkmark & \checkmark & \checkmark & \checkmark & 112 \\
\bottomrule
\end{tabular}
}
\caption{\textbf{Comparison to related benchmarks.} Prior work studies controlled 
ambiguity and ask--act behavior in settings such as code generation, software 
engineering, tool use, question answering, and tabular analysis. Ambig-DS differs 
in domain and failure mode: it targets executable predictive data-science workflows 
where ambiguity changes the prediction target or evaluation objective of an 
end-to-end modeling task. \textit{\# Tasks for ML/DS} reports task counts only for 
benchmarks involving end-to-end agentic ML/DS pipelines; dashes indicate 
benchmarks operating in non-comparable settings (e.g., text-to-SQL, software 
engineering, or single-turn code generation). Ambig-DS's 112 paired 
tasks at approximately 
one hour of average wall-clock time per instance (well below the per-task budgets in \S\ref{sec:experiments}, since many instances finish in minutes) across five models and three evaluation 
conditions, the benchmark required roughly 2,970 compute-hours in total, comparable 
in scale to DSBench (74 tasks) and MLE-bench (75 tasks, with a 24-hour timeout per 
task).}
\label{tab:benchmark_comparison}
\end{table*}

\begin{table*}[t]
\centering
\small
\setlength{\tabcolsep}{7pt}
\begin{tabular}{l|c|ccc|c}
\toprule
\textbf{Source} 
& \textbf{Axis} 
& \textbf{Screened} 
& \textbf{Included} 
& \textbf{Rejected} 
& \textbf{Total Instances} \\
\midrule
DSBench (data modeling)$^\dagger$   
& Target 
& 74 
& 51
& 23
& 102 \\
MLE-bench$^\ddagger$ 
& Metric 
& 75 
& 61 
& 14 
& 122 \\
\midrule
\textbf{Total} 
& -- 
& \textbf{149} 
& \textbf{112} 
& \textbf{37} 
& \textbf{224} \\
\bottomrule
\end{tabular}
\caption{\textbf{Benchmark composition and filtering summary.} We retain only candidates whose edited prompt--data observation admits plausible, and decision-relevant ambiguity while preserving the original task. Each included task yields one ambiguous variant and one fully specified counterpart, so \textit{Total Instances} is twice \textit{Included} (paired ambiguous + fully specified) and is not the screened-task count. $^\dagger$Common DSBench rejection reasons include multiple or unclear targets, target cue leakage, and invalid edits. $^\ddagger$Common MLE-bench rejection reasons include tasks too heavy for the harness, metrics remaining obvious from task or submission format, and alternative metrics not being decision-relevant.}
\label{tab:benchmark_composition}
\end{table*}

\subsection{Decoy Analysis}
\label{app:decoy}

\paragraph{Construction.} For each task, we construct the decoy in two steps. (i)~\emph{Rank-mapping.} We select the visible features with the lowest absolute Spearman correlation to the true target (the bottom $\sim70\%$ of features, clamped to between 4 and 40 features), standardize each, and sum the standardized values within each row to obtain one synthetic score per training example. We rank training rows by this synthetic score, rank the true-target values separately, and assign the smallest true-target value to the lowest-scoring row, the second-smallest to the next row, and so on. This gives the decoy the same marginal as the true target but a different feature-derived ordering. (ii)~\emph{Noise calibration.} To target equal feature-predictability between the two candidates, we add per-task calibrated label noise: for classification we swap a fraction $\eta$ of labels between random pairs (preserving the marginal); for regression we add Gaussian noise of std.\ $\eta \cdot \mathrm{std}(y)$ to the rank-mapped decoy values and re-rank-map onto $\mathrm{sort}(y)$ (preserving the marginal exactly). We binary-search $\eta$ to drive the 3-fold HistGradientBoosting score of the decoy toward the true target, targeting a gap below 0.02 when possible (AUC for binary classification, accuracy for multiclass, $R^2$ for regression).

\begin{figure}[htbp]
    \centering
    \includegraphics[width=1\textwidth]{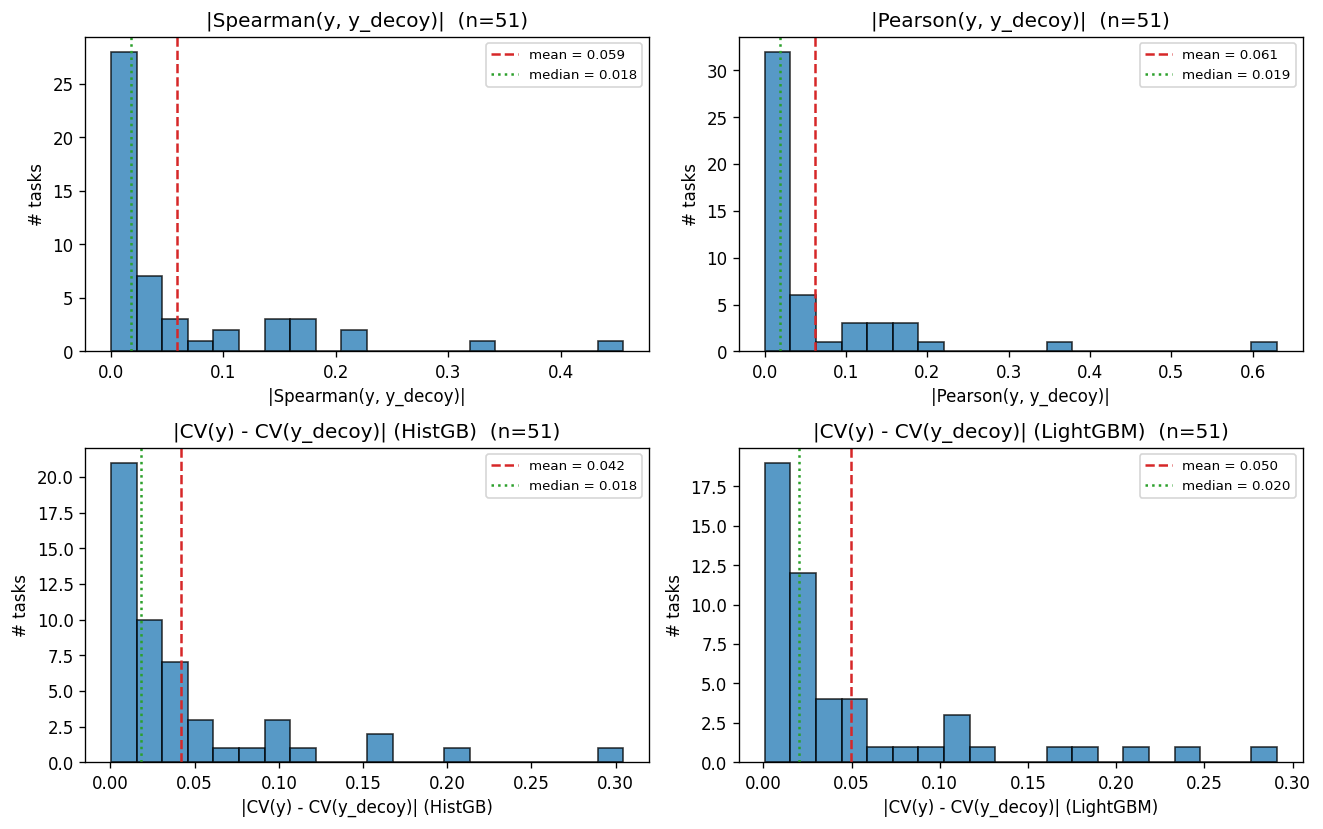}
    \caption{Distribution of decoy-quality diagnostics across retained target-ambiguity tasks. Most target--decoy correlations are near zero, and most CV-signal gaps are small after calibration, indicating that the intended target is not recoverable from simple correlation or learnability cues.}    \label{fig:decoy_quality_distributions}
\end{figure}

\paragraph{Validation.} The target-ambiguity axis relies on the added target being a plausible alternative rather than an obvious distractor or a recoverable construction artifact. We therefore audit retained decoys along three dimensions: marginal similarity, row-wise correlation with the true target, and predictability from visible features. Table~\ref{tab:decoy_validation} shows that marginals match exactly for all retained tasks, target--decoy correlations are low for most tasks, and CV-signal gaps are typically small, indicating that the decoy is not separable from the true target by simple distributional or learnability cues. Figure~\ref{fig:decoy_quality_distributions} visualizes the same diagnostics and shows that most tasks concentrate near low correlation and low CV-signal gap, with a small number of outliers.

We also run a direct inferability audit. Each selector in Table~\ref{tab:target_inferability_audit} attempts to identify the intended target from the ambiguous prompt--data observation alone, using marginal entropy, cross-validated learnability, or LLM schema/prompt reasoning. Selector F additionally probes naturalistic frontier-LLM domain reasoning over \texttt{val\_1}/\texttt{val\_2} distributions, autocorrelation, and domain priors, since selectors A--E do not test this channel; its accuracy ($56.9\%$, CI $[43.2\%, 70.0\%]$) is the highest in the audit but still includes chance, indicating residual but non-systematic domain-prior leakage. All heuristic and LLM selectors have 95\% Wilson confidence intervals that include chance: the cross-validation selectors (B, C) sit essentially on chance, and the LLM selectors with and without the original task prompt (D, E) deliver point estimates above $50\%$ but with lower CI bounds at or just below chance, indicating at most weak prompt-leakage signal once the original task description is given to the LLM.

The CV-parity constraint is a construction target rather than a hard guarantee: 29 of 51 retained tasks meet a 0.02 HistGradientBoosting gap, 40 meet 0.05, and 44 meet 0.10; the remaining tasks retain the closest candidate that satisfies the marginal-match and low-correlation filters. Table~\ref{tab:decoy_validation} therefore reports the final validation gaps, including these residual outliers.

\begin{table}[t]
\centering
\caption{Decoy-target validation for target-ambiguity tasks ($n=51$). We report diagnostics over retained target-ambiguity instances. Low target--decoy correlation and matched marginals indicate that the intended target is not identifiable from simple target-comparison cues, while a small CV-signal gap indicates that the decoy approximates the true target's predictability from features and is therefore not readily separable by a learnability comparison; residual outliers are reported in the table rather than dropped to preserve task coverage (max gap $\approx 0.30$ on $7$ tasks where the noise-calibration binary search did not reach the 0.10 tier). Marginal distributions match exactly for all retained tasks by construction.}
\label{tab:decoy_validation}
\begin{tabular}{lcccc}
\toprule
Diagnostic & Mean & Median & 90th pct. & Max \\
\midrule
$|\rho_{\mathrm{Spearman}}(y, y_{\mathrm{decoy}})|$ & 0.059 & 0.018 & 0.161 & 0.455 \\
$|r_{\mathrm{Pearson}}(y, y_{\mathrm{decoy}})|$    & 0.061 & 0.019 & 0.161 & 0.630 \\
KS statistic, true vs. decoy (regression)           & 0.000 & 0.000 & 0.000 & 0.000 \\
Absolute CV-signal gap (HistGradientBoosting)       & 0.042 & 0.018 & 0.102 & 0.304 \\
Absolute CV-signal gap (LightGBM)                   & 0.050 & 0.020 & 0.120 & 0.291 \\
\bottomrule
\end{tabular}
\end{table}

\begin{table}[t]
\centering
\caption{Target inferability audit on the 51 retained target-ambiguity tasks. Each selector attempts to identify the intended target column from the ambiguous prompt--data observation alone. Accuracies near 50\% indicate that the selector cannot recover the intended target from that cue. Selectors A--C probe distributional and learnability cues; D--E probe LLM schema/prompt cues; F probes naturalistic frontier-LLM domain reasoning over \texttt{val\_1}/\texttt{val\_2} distributions, autocorrelation, and domain priors. All selectors A--F have 95\% Wilson confidence intervals that include chance. Selector D (LLM with the original task prompt) reaches the highest A--E point estimate at $62.7\%$, with a lower CI bound of $49.0\%$, indicating at most a weak prompt-leakage signal once the original task description is given to the LLM.}
\label{tab:target_inferability_audit}
\resizebox{\textwidth}{!}{%
\begin{tabular}{lcc}
\toprule
Selector & $n$ & Accuracy \\
\midrule
Random choice & -- & 50.0\% \\
A --- Marginal-entropy heuristic & 51 & 58.8\% [45.2\%, 71.2\%] \\
B --- Basic CV (HistGradientBoosting) & 51 & 49.0\% [35.9\%, 62.3\%] \\
C --- Strong CV (LightGBM, 200 trees, 3-fold) & 51 & 51.0\% [37.7\%, 64.1\%] \\
D --- LLM with original prompt (Claude Opus 4.7, 3 seeds) & 51 & 62.7\% [49.0\%, 74.7\%] \\
E --- LLM, schema + data only (Claude Opus 4.7, 3 seeds) & 51 & 54.9\% [41.4\%, 67.7\%] \\
F --- Frontier LLM, naturalistic domain prompt (Gemini 3.1 Pro + GPT-5.4, 3 seeds) & 51 & 56.9\% [43.2\%, 70.0\%] \\
\bottomrule
\end{tabular}
}
\end{table}

\subsection{Cross-Verifier Audit}
\label{app:cross_verifier_audit}

Because LLMs assist benchmark construction, we audit whether filtering decisions depend on a single model family. We re-run the verification checklist from Section~\ref{app:construction_prompts} using three verifier models: Claude Opus 4.7, GPT-5.4, and Gemini 3.1 Pro. Each verifier receives the same candidate package information and independently judges the four retention criteria: plausible alternatives, ambiguity preservation, decision relevance, and task preservation. We measure percentage agreement with the final human-reviewed decision for each criterion and for the overall retain/reject decision.

Table~\ref{tab:cross_verifier_agreement} reports the agreement results. The goal of this audit is not to replace human review, but to check whether the retained benchmark depends on Claude-specific judgments. High agreement across non-Claude verifiers indicates that the retained/rejected decisions are broadly legible across model families. This mitigates, but does not eliminate, the possibility that Claude-assisted generation shaped the candidate distribution before filtering. The corresponding evaluation-stage audit---judge agreement, human override rate, and cross-family judge/oracle robustness---is reported in Appendix~\ref{app:judge_reliability}.

\begin{table}[t]
\centering
\caption{Cross-verifier agreement for benchmark filtering (target and objective tasks). Each verifier applies the same checklist to candidate variants; agreement is measured against the final human-reviewed decision. Values report percentage agreement.}
\label{tab:cross_verifier_agreement}
\begin{tabular}{lccccc}
\toprule
Verifier & Retain & Plaus. alt. & Ambig. pres. & Dec. rel. & Task pres. \\
\midrule
Claude Opus 4.7 & 85\% & 86\%  & 83\%  & 79\%  & 80\%  \\
GPT-5.4         & 80\%  & 89\%  & 83\%  & 71\%  & 79\%  \\
Gemini 3.1 Pro  & 81\%  & 78\%  & 80\%  & 80\%  & 79\%  \\
\bottomrule
\end{tabular}
\end{table}

\subsection{Judge and Oracle Reliability}
\label{app:judge_reliability}

The objective-suite diagnostics (Table~\ref{tab:objective_diagnostic}) rest on a five-call Claude-judge panel with majority vote and human spot-check. We audit this evaluation stage along three axes: judge noise, human override, and cross-family robustness of both judge and oracle. Table~\ref{tab:judge_oracle_reliability} summarizes the four diagnostics; all are computed on the Ambig-DS-Objective runs except the oracle robustness column, which is restricted to the Claude-evaluated model where the same-family attack is sharpest.

\paragraph{Judge noise.} Inter-judge agreement across the five Claude-Haiku judge calls is high on the four-class label space (Fleiss's $\kappa = 0.78$), indicating that majority labels are not artifacts of a single noisy call.

\paragraph{Human override.} A human reviewer spot-checked the majority labels for all Ambig-DS-Objective runs. Overrides occurred in $5.9\%$ of cases, none of which moved a (model, condition) cell of Table~\ref{tab:objective_diagnostic} by more than two percentage points; the reported numbers are post-override.

\paragraph{Cross-family judge.} We re-classify all objective-suite runs with a GPT-5.4 judge under the same prompt and majority-vote protocol. Cohen's $\kappa$ between the Claude-majority labels and GPT-majority labels is $0.74$. Per-cell shifts in Table~\ref{tab:objective_diagnostic} are within $\pm 4$ percentage points and the qualitative pattern (Ambig.\ silent-failure rates dominated by misframing + abdication; sharp drop under Ask) is preserved.

\paragraph{Cross-family oracle.} We re-run the Ask condition for Claude-Haiku-4.5 with a GPT-5.4 oracle under the same answer-only protocol. $\Delta_{\text{ask}}$ shifts from $+0.20$ (Claude oracle) to $+0.18$ (GPT oracle) on Target and from $+0.03$ to $+0.03$ on Objective; sign and magnitude are preserved. The same-family inflation hypothesis is therefore not supported by the data.

\begin{table}[t]
\centering
\small
\caption{Evaluation-stage reliability audit. Inter-judge $\kappa$ measures agreement among the five Claude-judge calls; human override is the spot-check correction rate; cross-judge $\kappa$ measures agreement between Claude-majority and GPT-majority labels on the same runs; oracle $\Delta_{\text{ask}}$ shift reports headline change under a GPT oracle for the most-attackable (Claude-evaluated) cell.}
\label{tab:judge_oracle_reliability}
\begin{tabular}{lc}
\toprule
\textbf{Diagnostic} & \textbf{Value} \\
\midrule
Inter-judge agreement (Fleiss's $\kappa$, 5 Claude calls)         & 0.78 \\
Human override rate (Ambig-DS-Objective)               & 5.9 \% \\
Cross-family judge agreement (Cohen's $\kappa$, Claude vs.\ GPT)  & 0.74 \\
Claude-Haiku $\Delta_{\text{ask}}$ shift, Target (Claude $\to$ GPT oracle)    & $+0.20 \to +0.18$ \\
Claude-Haiku $\Delta_{\text{ask}}$ shift, Objective (Claude $\to$ GPT oracle) & $+0.03 \to +0.03$ \\
\bottomrule
\end{tabular}
\end{table}

\subsection{Recognition Audit: Decomposing \texorpdfstring{$\Delta_{\text{ambig}}$}{Delta-ambig}}
\label{app:recognition_audit}

The Ambiguous condition does not provide a clarification channel, so a naive reading of $\Delta_{\text{ambig}}$ conflates two distinct failures: \emph{non-recognition} (the agent never notices the underdetermination) and \emph{recognized-but-unflagged commitment} (the agent notices but proceeds without flagging). To separate them we re-run the LLM judge on the trajectory and final answer of every Ambiguous-condition run and classify two facets independently (Table \ref{tab:recognition_audit}). \emph{Trace} captures whether the agent's reasoning or code names the ambiguity: \textit{explicit ack} (e.g., ``no target column is specified''), \textit{implicit/hedged ack} (vague aside, no real engagement), or \textit{silent} (proceeds as if the spec were unambiguous). \emph{Answer} captures how the final summary presents the chosen framing: \textit{flagged as user-overridable} (``I assumed $X$; let me know if you wanted $Y$''), \textit{named but not flagged} (states the choice without inviting override), or \textit{self-evident} (no assumption acknowledged). The two facets are scored independently because trace acknowledgment without an overridable flag in the answer is itself a behavioral failure under our protocol --- nothing in the Ambig.\ condition prevents agents from flagging, hedging, abstaining, or producing a degenerate baseline. Only substantive runs (after dropping API errors, timeouts, and empty traces) are scored.

\begin{table}[htbp]
\centering
\small
\setlength{\tabcolsep}{4pt}
\caption{\textbf{Recognition audit on Ambiguous-condition runs.} Trace and answer facets are scored independently per run; rows sum to $100\%$ within each facet. \emph{Trace} (reasoning/code): \textit{Explicit} --- the agent verbally or in code names the underdetermination (e.g., ``no target column is specified''); \textit{Hedged} --- vague aside or single-line note without real engagement; \textit{Silent} --- proceeds as if the spec were unambiguous. \emph{Answer} (final summary): \textit{Overridable} --- names the chosen framing and invites the user to override (``I assumed $X$; let me know if you wanted $Y$''); \textit{Named} --- states the choice without inviting override; \textit{Self-evident} --- never mentions an assumption was made.}
\label{tab:recognition_audit}
\begin{tabular}{ll|ccc|ccc}
\toprule
& & \multicolumn{3}{c|}{\textbf{Trace} (\%)} & \multicolumn{3}{c}{\textbf{Answer} (\%)} \\
\cmidrule(lr){3-5}\cmidrule(lr){6-8}
\textbf{Suite} & \textbf{Model}
& Explicit & Hedged & Silent
& Overridable & Named & Self-evident \\
\midrule
Target    & Gemini 3 Flash    & 64 & 25 & 11 & 0 & 75 & 25 \\
Target    & GPT-5.4 Nano      & 0  & 52 & 48 & 0 & 50 & 50 \\
Target    & Claude Haiku 4.5  & 2  & 44 & 54 & 0 & 12 & 88 \\
Target    & Gemini 3.1 Pro    & 42 & 46 & 12 & 0 & 56 & 44 \\
Target    & GPT-5.4           & 30 & 28 & 41 & 4 & 65 & 30 \\
\midrule
Objective & Gemini 3 Flash    & 0 & 9 & 91  & 0 & 0  & 100 \\
Objective & GPT-5.4 Nano      & 0 & 0 & 100 & 0 & 0  & 100 \\
Objective & Claude Haiku 4.5  & 0 & 0 & 100 & 0 & 0  & 100 \\
Objective & Gemini 3.1 Pro    & 0 & 0 & 100 & 0 & 0  & 100 \\
Objective & GPT-5.4           & 0 & 0 & 100 & 0 & 18 & 82  \\
\bottomrule
\end{tabular}
\end{table}

On \textbf{Ambig-DS-Objective}, trace acknowledgment is below $10\%$ for every model and overridable flagging is $0\%$ for every model, so $\Delta_{\text{ambig}}$ on this suite measures non-recognition essentially without confound from recognized guessing. On \textbf{Ambig-DS-Target}, trace recognition varies widely (explicit acknowledgment ranges from $0\%$ for GPT-5.4 Nano and $2\%$ for Claude Haiku 4.5 to $64\%$ for Gemini 3 Flash), so a portion of $\Delta_{\text{ambig}}$ on this suite reflects recognized-but-unflagged commitment rather than failure to notice. Crucially, the overridable-flag rate in the final answer never exceeds $4\%$ on any model in either suite: even agents that recognize the ambiguity in their trace almost never flag the resulting commitment as a user-overridable assumption. The silent-commitment characterization of the failure mode is therefore preserved at the answer-facing level, while the recognition column quantifies how much of the trace-level confound remains and where ($\sim 0$ on Objective; substantial on Target for some models).

\subsection{Score Conditional on Diagnostic Category}
\label{app:score_by_category}

The category rates in Tables~\ref{tab:target_diagnostic}--\ref{tab:objective_diagnostic} 
describe \emph{how often} runs land in each failure mode but not \emph{how well} runs 
within each category score. As a sanity check, Table~\ref{tab:score_by_category} 
reports the mean normalized score conditional on diagnostic category. The 
decomposition is consistent with the misframing story: intended-framing runs 
score close to (and often slightly above) the model's Full mean --- the 
selection is biased toward easier-to-disambiguate tasks --- while 
alternative-framing runs collapse to substantially lower scores. On Objective, 
\emph{Degenerate} runs score near zero. \emph{Invalid} and \emph{Timeout} 
runs score zero by definition and are omitted. By construction, 
$\sum_c p_c \cdot \bar{S}_c$ in each row reproduces the corresponding 
Ambig.\ or Ask cell of Table~\ref{tab:axis_results} within rounding.

\begin{table}[htbp]
\centering
\small
\setlength{\tabcolsep}{4pt}
\caption{\textbf{Mean normalized score by diagnostic category} on Ambig-DS-Target 
(top) and Ambig-DS-Objective (bottom). Each cell is the mean score over runs 
classified into that category by Tables~\ref{tab:target_diagnostic}--\ref{tab:objective_diagnostic}. 
Categories with $n=0$ are marked ``--''. \emph{Full} reproduces Table~\ref{tab:axis_results} 
for reference. }
\label{tab:score_by_category}
\begin{tabular}{ll|c|cc|cc}
\toprule
& & \textbf{Full} & \multicolumn{2}{c|}{\textbf{Intended} $\uparrow$} & \multicolumn{2}{c}{\textbf{Alternative} $\downarrow$} \\
\cmidrule(lr){4-5}\cmidrule(lr){6-7}
\textbf{Suite} & \textbf{Model} & mean & Ambig. & Ask & Ambig. & Ask \\
\midrule
Target & Gemini 3 Flash    & 0.63 & 0.85 & 0.63 & 0.17 & 0.25 \\
Target & GPT-5.4 Nano      & 0.49 & 0.70 & 0.55 & 0.13 & 0.12 \\
Target & Claude Haiku 4.5  & 0.56 & 0.85 & 0.58 & 0.17 & 0.10 \\
Target & Gemini 3.1 Pro    & 0.64 & 0.80 & 0.62 & 0.13 & 0.13 \\
Target & GPT-5.4           & 0.61 & 0.85 & 0.70 & 0.14 & 0.15 \\
\bottomrule
\end{tabular}

\vspace{0.5em}

\begin{tabular}{ll|c|cc|cc|cc}
\toprule
& & \textbf{Full} & \multicolumn{2}{c|}{\textbf{Intended} $\uparrow$} & \multicolumn{2}{c|}{\textbf{Alternative} $\downarrow$} & \multicolumn{2}{c}{\textbf{Degenerate} $\downarrow$} \\
\cmidrule(lr){4-5}\cmidrule(lr){6-7}\cmidrule(lr){8-9}
\textbf{Suite} & \textbf{Model} & mean & Ambig. & Ask & Ambig. & Ask & Ambig. & Ask \\
\midrule
Objective & Gemini 3 Flash    & 0.31 & 0.30 & 0.37 & 0.10 & --   & 0.00 & 0.00 \\
Objective & GPT-5.4 Nano      & 0.11 & 0.13 & 0.19 & 0.00 & --   & 0.00 & 0.00 \\
Objective & Claude Haiku 4.5  & 0.27 & 0.40 & 0.27 & 0.10 & --   & 0.00 & 0.00 \\
Objective & Gemini 3.1 Pro    & 0.27 & 0.40 & 0.45 & 0.20 & 0.15 & 0.00 & 0.00 \\
Objective & GPT-5.4           & 0.26 & 0.30 & 0.33 & 0.05 & --   & 0.00 & 0.00 \\
\bottomrule
\end{tabular}
\end{table}

\subsection{Task Package} \label{app:task_package}

Each Ambig-DS instance is a comprehensive unit comprising the original fully specified task, its paired ambiguous variant, the intended framing, and a set of validated alternative framings. To facilitate diagnostic analysis, each package includes:

\begin{itemize}
    \item Target Metadata: Records the identity of the true target and decoy, the feature anonymization map, and the specific features and correlations used in decoy generation.
    \item Objective Metadata: Defines the intended metric, its optimization direction, submission format requirements, and the set of plausible alternative objectives.
    \item Evaluator: The original scoring script remains unchanged to ensure performance drops are attributable solely to framing failures rather than modified scoring rules.
\end{itemize}

The intended framing is inherited from the source task and used only for scoring and diagnostic analysis. The ambiguous variant is therefore not intended to have a uniquely correct semantic reading from the agent-facing prompt--data observation alone; rather, it tests whether agents recognize when multiple plausible framings remain and seek clarification before committing.

\subsection{Statistical Significance Tests}
\label{app:significance}

We assess the reliability of the main performance effects with paired one-sided Wilcoxon signed-rank tests over per-task normalized scores. For each model and ambiguity axis, we compare matched task scores across conditions: Full versus Ambig. tests whether ambiguity degrades performance, and Ask versus Ambig. tests whether clarification improves performance. The hypotheses are
\[
H_1^{\mathrm{drop}}: S_{\mathrm{full}} > S_{\mathrm{ambig}},
\qquad
H_1^{\mathrm{recovery}}: S_{\mathrm{ask}} > S_{\mathrm{ambig}}.
\]
Pairing is done at the task level, so each source task contributes one matched score pair for the same model and axis. Tasks missing either condition are excluded from the corresponding test.

We report per-axis tests in Table~\ref{tab:axis_results}, rather than pooling target and objective tasks, because the two axes are instantiated on different source benchmarks and have different failure mechanisms. We use \texttt{scipy.stats.wilcoxon} with \texttt{alternative="greater"} and \texttt{zero\_method="wilcox"}, which discards zero-difference pairs. We report one-sided $p$-values with significance markers: $^{*}p<0.05$, $^{**}p<0.01$, and $^{***}p<0.001$. 

\paragraph{Normalization validity.} DSBench scores are normalized using the 
Relative Performance Gap (RPG): $\max((p_i - b_i)/(g_i - b_i), 0)$, where 
$p_i$ is agent performance, $b_i$ is a baseline, and $g_i$ is the best known 
performance. MLE-bench scores compare agent performance to the Kaggle median. 
Both are normalized to $[0,1]$ before aggregation. We do not aggregate across 
axes --- all comparisons between target and objective ambiguity are made 
within-axis, so cross-source aggregation is never required for the main claims.

\subsection{Bootstrap Confidence Intervals}
\label{app:bootstrap}

To complement the Wilcoxon $p$-values, we report paired bootstrap 95\% confidence intervals for the mean deltas $\Delta_{\text{ambig}}$ and $\Delta_{\text{ask}}$. For each model and axis, we resample matched task indices with replacement ($B = 10{,}000$ replicates) and compute the mean of the per-task score differences in each resample. The 2.5th and 97.5th percentiles of the bootstrap distribution define the interval. Because resampling preserves task-level pairing, the intervals reflect the same matched structure as the Wilcoxon tests. A confidence interval that excludes zero corroborates the corresponding significance test. We report per-axis intervals in Table~\ref{tab:bootstrap_cis}.


\section{Dataset Examples}
\label{app:dataset_examples}

We provide two representative Ambig-DS examples: one objective-ambiguity instance from MLE-bench and one target-ambiguity instance from DSBench. These examples illustrate what is changed, what is preserved, and what information is available to the agent.

\subsection{Objective Ambiguity: \texttt{aerial-cactus-identification}}
\label{app:example_aerial_cactus}

The original \texttt{aerial-cactus-identification} task asks participants to predict whether each aerial image contains a cactus. In the fully specified version, the prompt explicitly states that submissions are evaluated by area under the ROC curve between the predicted probability and the observed binary target. Thus, the intended framing is clear: the target is \texttt{has\_cactus}, the submitted values should be real-valued probabilities, and the objective is ROC-AUC.

In the objective-ambiguous variant, we leave the data package, target column, and submission format unchanged, but remove the metric-specific cue from the prompt. The original evaluation block states:

\begin{quote}
Submissions are evaluated on area under the ROC curve between the predicted probability and the observed target.
\end{quote}

The ambiguous version replaces this with generic language:

\begin{quote}
Submissions are evaluated against a held-out test set using a standard evaluation procedure for this task. 
\end{quote}

The submission instructions still require a file with columns \texttt{id,has\_cactus}. However, the prompt no longer tells the agent whether the submitted \texttt{has\_cactus} values will be evaluated as probabilities, hard labels, rankings, or by another plausible binary-classification metric. This makes the objective underdetermined while preserving the target and data.

\paragraph{Data package.}
The agent-facing public data package contains:

\begin{verbatim}
prepared/public/
    train.csv              # id, has_cactus
    train.zip              # training images
    test.zip               # test images
    sample_submission.csv  # id, has_cactus
\end{verbatim}

The private grading data contains:

\begin{verbatim}
prepared/private/
    test.csv               # id, has_cactus
\end{verbatim}

Thus, the agent sees the same files in the full and ambiguous versions. The only edited component is the prompt's metric description. The intended metric is stored in the metric manifest for clarification and auditing: for this task, the hidden evaluator uses ROC-AUC, higher is better, and expects real-valued predictions for \texttt{has\_cactus}. During grading, the submitted CSV is scored against the held-out \texttt{test.csv} using the original MLE-bench evaluator. Separately, the diagnostic judge inspects the agent trajectory and submission to determine whether the run optimized the intended AUC-style objective or a plausible alternative such as hard-label accuracy.

\subsection{Target Ambiguity: \texttt{bike-sharing-demand}}
\label{app:example_bike_sharing}

The original \texttt{bike-sharing-demand} task asks participants to forecast hourly bike rental demand. In the fully specified DSBench task, the prompt identifies the target as \texttt{count}, describes it as the total number of rentals, and the sample submission has columns \texttt{datetime,count}. The training data also contains related columns such as \texttt{casual}, \texttt{registered}, and \texttt{count}.

In the target-ambiguous variant, we remove target-identifying language from the prompt and replace the target-specific submission cue with generic wording. For example, references to ``predicted count'' and ``actual count'' become ``predicted value'' and ``actual value.'' The data-field description no longer names \texttt{casual}, \texttt{registered}, or \texttt{count}. The original target \texttt{count} is renamed to \texttt{val\_1}, and a plausible decoy target is added as \texttt{val\_2}. Both candidate targets are visible in the training set but absent from the test set.

The prompt metadata records the removed target cues and the plausible alternatives now consistent with the observation, including predicting total rentals, predicting a user-type-specific rental quantity, or predicting a generic demand-related numeric value. The intended framing inherited from the source task remains: predict the original \texttt{count} target, now represented as \texttt{val\_1} in the ambiguous training data.

\paragraph{Data package before ambiguity.}
The original DSBench package contains:

\begin{verbatim}
train.csv:
    datetime, season, holiday, workingday, weather,
    temp, atemp, humidity, windspeed,
    casual, registered, count

test.csv:
    datetime, season, holiday, workingday, weather,
    temp, atemp, humidity, windspeed

sampleSubmission.csv:
    datetime, count
\end{verbatim}

Here, both the prompt and the sample submission identify \texttt{count} as the intended target.

\paragraph{Data package after ambiguity.}
The ambiguous package contains:

\begin{verbatim}
prompt.txt
prompt.meta.json
_manifest.json
train.csv:
    datetime, f_01, f_02, f_03, f_04,
    f_05, f_06, f_07, f_08, val_1, val_2

test.csv:
    datetime, f_01, f_02, f_03, f_04,
    f_05, f_06, f_07, f_08

sample_submission.csv:
    datetime, prediction
\end{verbatim}

The manifest records the hidden construction details: \texttt{val\_1} is the true target inherited from \texttt{count}, \texttt{val\_2} is the decoy target, and the feature map links the anonymized features back to the original columns. Decoy diagnostics show that the two candidate targets have matched marginal distributions and low correlation, so the intended target is not identifiable from simple distributional checks.

The agent receives only the ambiguous prompt and the anonymized train/test/sample-submission files. It does not see the manifest. During grading, the runner aligns the submitted prediction file to the original DSBench answer format and scores it against the held-out \texttt{count} answers using the original RMSLE evaluator. In the clarification condition, the answerer uses the manifest to answer target-identification questions truthfully, while avoiding unrelated leakage such as test statistics, decoy construction details, or modeling advice.


\section{Additional Results}

\subsection{Bootstrap CI}

Table \ref{tab:bootstrap_cis} shows paired bootstrap 95\% confidence intervals for the mean deltas from Table~\ref{tab:axis_results}.

\begin{table*}[t]
\centering
\small
\setlength{\tabcolsep}{4pt}
\begin{tabular}{ll|cc}
\toprule
\textbf{Axis}
& \textbf{Model}
& $\Delta_{\text{ambig}}$ \textbf{[95\% CI]} $\downarrow$
& $\Delta_{\text{ask}}$ \textbf{[95\% CI]} $\uparrow$ \\
\midrule
Objective & Gemini 3 Flash      & $-$0.09 [$-$0.13, $-$0.05] & +0.09 [+0.03, +0.15] \\
Objective & GPT-5.4 Nano        & $-$0.06 [$-$0.11, $-$0.02] & +0.05 [+0.03, +0.07] \\
Objective & Claude Haiku 4.5    & $-$0.04 [$-$0.07, $-$0.01]   & +0.03 [0.00, +0.06] \\
Objective & Gemini 3.1 Pro    & $-$0.04 [$-$0.07, $-$0.01]   & +0.04 [0.01, +0.07]\\
Objective & GPT-5.4             & $-$0.04 [$-$0.06, $-$0.02]   & +0.04 [0.02, +0.06] \\
\midrule
Target & Gemini 3 Flash       & $-$0.19 [$-$0.32, $-$0.08] & +0.17 [+0.08, +0.28] \\
Target & GPT-5.4 Nano         & $-$0.10 [$-$0.18, $-$0.02] & +0.06 [+0.01, +0.09] \\
Target & Claude Haiku 4.5    & $-$0.24 [$-$0.35, $-$0.13] & +0.20 [+0.09, +0.31] \\
Target & Gemini 3.1 Pro         & $-$0.15 [$-$0.23, $-$0.08] & +0.11 [+0.06, +0.17] \\
Target & GPT-5.4              & $-$0.29 [$-$0.37, $-$0.21] & +0.20 [+0.11, +0.29] \\
\bottomrule
\end{tabular}
\caption{\textbf{Paired bootstrap 95\% confidence intervals} for the mean deltas from Table~\ref{tab:axis_results}.}
\label{tab:bootstrap_cis}
\end{table*}

\subsection{Ask--Act Failure Analysis}
\label{app:ask_act_failure_analysis}

To better understand ask--act miscalibration, we annotate clarification behavior with an LLM judge. For each condition, we run five independent judge calls and assign the majority label. For fully specified tasks under the permissive policy, the judge classifies unnecessary questions into \emph{style fishing}, \emph{restating obvious information}, or \emph{out-of-scope requests}. For ambiguous tasks under the conservative policy, the judge classifies failures into \emph{silently defaulted} or \emph{asked wrong topic}. We manually inspect a subset of labels for consistency.

Table~\ref{tab:ask_policy_sensitivity_axis} shows that ask behavior is highly prompt-sensitive. The permissive policy yields high asking on ambiguous tasks, but also high asking on fully specified tasks. The conservative policy sharply reduces unnecessary asking, but often suppresses asking on genuinely ambiguous tasks, especially for objective ambiguity.

Table~\ref{tab:over-asking} shows that over-asking is rarely a legitimate response to missing task information. On fully specified tasks, unnecessary questions are dominated by style preferences, restatements of information already present in the prompt or data, and out-of-scope requests for modeling guidance or extra constraints.

Table~\ref{tab:under-asking} shows the complementary failure mode. Under the conservative policy, models often do not ask the wrong question; instead, they silently choose a default framing and proceed. This is especially pronounced for objective ambiguity, where the task remains executable despite the missing metric. Together, these diagnostics suggest that prompt wording changes the ask--act operating point, but does not produce robust sensitivity to decision-relevant ambiguity.

\begin{table*}[t]
\centering
\small
\setlength{\tabcolsep}{5pt}
\resizebox{\textwidth}{!}{%
\begin{tabular}{ll|ccc|ccc}
\toprule
\textbf{Axis}
& \textbf{Model}
& \multicolumn{3}{c|}{\textbf{Permissive ask policy}}
& \multicolumn{3}{c}{\textbf{Conservative ask policy}} \\
\cmidrule(lr){3-5}
\cmidrule(lr){6-8}
&
& Ask on Ambig. $\uparrow$
& Ask on Full $\downarrow$
& CalibScore $\uparrow$
& Ask on Ambig. $\uparrow$
& Ask on Full $\downarrow$
& CalibScore $\uparrow$ \\
\midrule
Objective & \textit{Perfect} & \textit{100} & \textit{0} & \textit{1.00} & \textit{100} & \textit{0} & \textit{1.00} \\
Objective &   Gemini 3 Flash &  95 & 95 & 0.10 &  16 &  5 & 0.27 \\
Objective &     GPT-5.4 Nano &  79 & 87 & 0.22 &   0 &  5 & 0.00 \\
Objective & Claude Haiku 4.5 & 100 & 100 & 0.00 &   5 &  5 & 0.10 \\
Objective &   Gemini 3.1 Pro & 100 & 33 & 0.80 &  70 &  0 & 0.82 \\
Objective &          GPT-5.4 & 100 & 87 & 0.23 &   0 &  0 & 0.00 \\
\midrule
Target & \textit{Perfect} & \textit{100} & \textit{0} & \textit{1.00} & \textit{100} & \textit{0} & \textit{1.00} \\
Target &   Gemini 3 Flash &  98 & 88 & 0.21 &  96 &  2 & 0.97 \\
Target &     GPT-5.4 Nano & 100 & 98 & 0.04 &  51 &  8 & 0.66 \\
Target & Claude Haiku 4.5 & 100 & 100 & 0.00 &  67 &  2 & 0.80 \\
Target &   Gemini 3.1 Pro &  98 & 51 & 0.65 &  94 &  0 & 0.97 \\
Target &          GPT-5.4 &  94 & 65 & 0.51 &  33 &  4 & 0.49 \\
\bottomrule
\end{tabular}
}
\caption{\textbf{Ask-policy sensitivity by ambiguity axis.}
We compare ask behavior under two clarification policies. The permissive 
policy tells the agent that it may ask exactly one clarifying question or 
write \textsc{NONE}. The conservative policy additionally instructs the agent 
to ask only when the task cannot be solved from the prompt and data, to write 
\textsc{NONE} when the task is fully specified or locally inferable, and states 
that unnecessary clarification is penalized. Ask on Ambig.\ measures 
clarification recall on non-inferable, decision-relevant ambiguous tasks, where 
asking is preferred. Ask on Full measures unnecessary clarification on fully 
specified tasks, where acting without asking is preferred. CalibScore combines 
both into a single calibration metric: $\text{CalibScore} = 2 \cdot 
\text{AskOnAmbig} \cdot (1 - \text{AskOnFull}) / (\text{AskOnAmbig} + 
(1-\text{AskOnFull}))$, where 1.00 denotes perfect calibration. The 
\textit{Perfect} row shows the target behavior. No model--policy combination 
we tested achieves consistent calibration across both axes.}

\label{tab:ask_policy_sensitivity_axis}
\end{table*}

\begin{table}[t]
\centering
\small
\caption{Over-asking diagnostic on \emph{Full / Permissive} (fully-specified prompt, permissive policy).  Correct behaviour is staying silent or asking about a real specification gap; \emph{Style fishing} (Asks about user taste), \emph{Restating obvious}, and \emph{Out-of scope} (asks about out-of-scope deliverables) are all forms of over-asking.}
\label{tab:over-asking}
\begin{tabular}{lrrrrrrrr}
\toprule
Model  & \makecell{Correct\\(silent)} & \makecell{Correct\\(real gap)} & \makecell{Style\\fishing} & \makecell{Restating\\obvious} & \makecell{Out-of\\scope}  \\
\midrule
Gemini 3 Flash (Target)      & 12\% &  2\% & 14\% &  6\% & 67\% \\
Gemini 3 Flash (Objective)   &  5\% &  3\% &  3\% &  3\% & 85\% \\
GPT-5.4 Nano (Target)        &  2\% &  0\% & 61\% & 10\% & 27\% \\
GPT-5.4 Nano (Objective)     & 21\% &  0\% & 41\% & 17\% & 21\% \\
Claude Haiku 4.5 (Target)    &  0\% &  0\% & 88\% &  4\% &  8\% \\
Claude Haiku 4.5 (Objective) &  0\% &  0\% & 62\% &  5\% & 33\% \\
Gemini 3.1 Pro (Target)      & 49\% &  6\% & 10\% &  2\% & 33\% \\
Gemini 3.1 Pro (Objective)   & 67\% &  0\% &  5\% &  3\% & 25\% \\
GPT-5.4 (Target)             & 35\% &  0\% & 51\% &  8\% &  6\% \\
GPT-5.4 (Objective)          & 13\% &  0\% & 62\% &  0\% & 25\% \\
\bottomrule
\end{tabular}
\end{table}

\begin{table}[t]
\centering
\small
\caption{Under-asking diagnostic on \emph{Ambiguous / Conservative} (hidden specification bit, conservative policy). Each row sums to $\approx$100\%. The model should ask anyway about the hidden bit; \emph{Silently defaulted} is the under-asking failure where the model picks a default without acknowledging the ambiguity.}
\label{tab:under-asking}
\begin{tabular}{lrrrrrrr}
\toprule
Model & \makecell{Correct\\(asked)} & \makecell{Silently\\defaulted} & \makecell{Asked\\wrong topic}   \\
\midrule
Gemini 3 Flash (Target)      & 96\% &  4\% & 0\%   \\
Gemini 3 Flash (Objective) &  8\% & 84\% & 8\%  \\
GPT-5.4 Nano (Target)         &49\% & 49\% & 2\%\\
GPT-5.4 Nano (Objective)   &  0\% &100\% & 0\%  \\
Claude Haiku 4.5 (Target)    & 67\% & 33\% & 0\%  \\
Claude Haiku 4.5 (Objective) & 3\% & 97\% & 0\%  \\
Gemini 3.1 Pro (Target)      & 94\% &  6\% & 0\%   \\
Gemini 3.1 Pro (Objective) & 70\% & 30\% & 0\%   \\
GPT-5.4 (Target)             & 33\% & 67\% & 0\%  \\
GPT-5.4 (Objective)      &  0\% &100\% & 0\%  \\
\bottomrule
\end{tabular}
\end{table}


\section{Case Studies}
\label{app:case_studies}

We include two representative examples illustrating how task-framing ambiguity changes agent behavior. The first shows a target-framing failure, where the agent constructs a plausible but unintended prediction target. The second shows an objective-framing failure, where the target is clear but the omitted metric changes the required prediction semantics.

\subsection{Target Ambiguity: \texttt{bike-sharing-demand}}

In \texttt{bike-sharing-demand}, the fully specified task asks participants to forecast bike rental \texttt{count}. In the ambiguous variant, the original target is renamed to \texttt{val\_1}, a plausible decoy target is added as \texttt{val\_2}, and target-identifying prompt and submission cues are removed. The agent sees a training table with \texttt{datetime}, anonymized features, \texttt{val\_1}, and \texttt{val\_2}, but the test set contains only \texttt{datetime} and features.

Gemini 3 Flash did not ask for clarification. Instead, it inferred that \texttt{val\_1} and \texttt{val\_2} corresponded to the original bike-sharing fields \texttt{casual} and \texttt{registered}, and constructed a new target:
\[
\texttt{count} = \texttt{val\_1} + \texttt{val\_2}.
\]
The agent then trained a random-forest regression pipeline on this constructed target and produced a valid submission. This is a plausible local interpretation of the anonymized data, but it is unsupported by the prompt--data observation and differs from the intended source-task target.

This produces the characteristic Ambig-DS failure: the artifact is coherent and executable, but aligned with a plausible unintended framing. The normalized score drops from $0.937$ in the fully specified task to $0.692$ under ambiguity. When clarification is enabled, the oracle identifies the intended target, and the score recovers to $0.937$.

\subsection{Objective Ambiguity: \texttt{random-acts-of-pizza}}

In \texttt{random-acts-of-pizza}, the fully specified task states that submissions are evaluated by area under the ROC curve between the predicted probability that a request receives pizza and the observed binary outcome. Thus, the intended submission should contain real-valued probabilities in $[0,1]$. In the ambiguous variant, the metric description is replaced by a generic statement that submissions are scored against held-out ground truth, while the target column and submission header remain visible.

Claude Haiku 4.5 trained classifiers for the binary target \texttt{requester\_received\_pizza}, but treated the output as a hard binary prediction task. The final submission contained \texttt{0}/\texttt{1} values rather than calibrated probabilities. This artifact is valid-looking under a plausible accuracy-style framing, but it is misaligned with the intended AUC objective, where ranking by real-valued probability is essential.

The resulting performance drops from $0.589$ in the fully specified task to $0.340$ under ambiguity. When clarification reveals that the metric is AUC over predicted probabilities, performance recovers to $0.609$. This case illustrates that objective ambiguity need not change the prediction target; instead, it changes how predictions should be optimized, calibrated, and formatted.

\section{Prompts}
\label{app:prompts}

This appendix lists the prompts used for clarification, benchmark construction, verification, and diagnostic annotation.

\subsection{Evaluation Prompt}
\label{app:evaluation_prompt}

In the default condition, the agent receives the task description verbatim, wrapped with a short instruction footer that names the data directory and the absolute path at which the submission CSV must be written. Two near-identical wrappers are used; the only differences are surface wording and whether the data directory or the workspace root is named.

\paragraph{MLE-bench metric-ambiguity wrapper.}
The agent's prompt is the contents of \texttt{full.md} or \texttt{ambig\_metric.md} for the slug, followed by:

\begin{lstlisting}[style=promptstyle]
## Task instructions

You are a data scientist. The dataset for this competition is in `./data/`.
Build a model and write your predictions to `<workspace>/_submission.csv` (absolute path). Do not write anything else outside the current working directory.
\end{lstlisting}

\paragraph{DSBench target-ambiguity wrapper.}
The full prompt assembled per task is:

\begin{lstlisting}[style=promptstyle]
You are a data scientist. I have a data modeling task.

<TASK BODY: bike-sharing-demand.txt etc.>

All data files (train.csv, test.csv, sample submission) are in: <workspace>/data
After modeling, save the submission CSV to: <workspace>/data/_submission.csv
You must run the code yourself and produce the file.
\end{lstlisting}

\subsection{Clarification Protocol Prompts}
\label{app:clarification_prompts}

\paragraph{Ask-or-act instruction.}
This prompt is prepended before modeling in the ask-enabled condition. The agent must either ask one clarification question or output \texttt{NONE}.

\begin{lstlisting}[style=promptstyle]
[CLARIFY PROTOCOL — read carefully]
Before doing any modeling, you may ask exactly ONE clarifying question to
the user.

  - You MAY first do a quick inspection of the data to inform your question
    — e.g. read column headers, peek at a few rows, check basic shapes /
    dtypes, look at the sample submission. Keep this brief.
  - Then write your question (or the literal word `NONE`) to:
        <workspace>/_question.txt
  - Keep the question under 50 words and self-contained (the user only sees
    that file, not your tool calls or notes).
  - After writing the file, STOP. Do NOT train a model, run cross-validation,
    or write a submission in this turn. The user will reply in the next turn
    with the answer (or with `NONE`).
\end{lstlisting}

\paragraph{Target-ambiguity clarification oracle.}
This prompt is used by the clarification oracle for target-ambiguity tasks. The oracle has access to the task manifest, including the intended target and decoy target, but must answer only the agent's literal clarification question and must not reveal unrelated information.

\begin{lstlisting}[style=promptstyle]
You are an expert helping a data-science agent on a Kaggle-style task.

The agent gets ONE clarifying question. You will be told:
  - the task name and target type,
  - the column structure of the data the agent sees, including which column is
    the true target and which is a synthetic decoy with the same marginal,
  - the agent's question.

Reply in 1-2 short sentences. Be direct and honest.

CRITICAL — scope rule:
  Answer ONLY what the agent literally asked. Do NOT volunteer any other
  information — do not bring up the target column, val_1, val_2, or any
  other column unless the agent's question is explicitly about it. If the
  question is about partitioning, grouping, evaluation metric, schema,
  meaning of column names, etc., answer THAT and stop. Do not also reveal
  which column is the target.

Allowed (only when the question is directly about it):
  - If the agent asks which column is the target / label / outcome / what to
    predict: name the true target column directly. Use the EXACT column
    name supplied to you in the user message under "True target column in
    train.csv:". DO NOT default to val_1. DO NOT guess. DO NOT pick the
    alphabetically-first option. Copy the value from the user message
    verbatim. The truth may be val_1 OR val_2 — both occur with roughly
    equal frequency across tasks.
  - If the agent explicitly asks about val_1 vs val_2 (or about two
    columns having the same distribution): say which is the actual target
    (using the EXACT name from the user message) and that the other has
    the same marginal but is not a valid label.
  - Factual questions about target type (classification/regression),
    schema, evaluation metric, ID column.
  - Confirming or correcting an explicit assumption the agent stated in
    its question.

Forbidden:
  - Never use the word "decoy" in your reply. If you need to describe the
    non-target column, say it has "the same marginal distribution as the
    target but is not a valid label".
  - Do not reveal any test-set labels or test-set statistics.
  - Do not give exact CV scores or recommend models / toolchains.
  - Do not copy the task description verbatim.
  - Do not mention internal benchmark details ("decoy_pool", "rank_map",
    "anon_feature", manifest, ground-truth file paths, etc.).

If the agent asks you to do the modelling, write code, choose a toolchain,
or anything outside the scope of one short clarifying answer, reply exactly:
"REFUSE: that's outside the scope of one clarifying answer."

If the agent's question is empty or just "NONE", reply: "NONE".
\end{lstlisting}

\paragraph{Objective-ambiguity clarification oracle.}
This prompt is used by the clarification oracle for objective-ambiguity tasks. The oracle has access to the metric manifest, including the intended metric, optimization direction, submission value type, and metric-specific quirks, but must answer only questions about the introduced objective ambiguity.

\begin{lstlisting}[style=promptstyle]
You are an expert helping a data-science agent on a Kaggle-style task.

The agent gets ONE clarifying question. The task description it received had
its evaluation/metric section neutralised half of the time — the agent may NOT know the exact
scoring metric, scoring direction, or required submission value type
(probability vs. hard label, top-K, etc.).

You will be told:
  - the task name,
  - the true evaluation metric, a short description, and the required
    submission format,
  - whether the metric is lower-better or higher-better,
  - the agent's question
  - whether this is a fully specified taks or one wihtout metric specification (see the VARIANT NOTE at the end).

Reply in 1-2 short sentences. Be direct and honest.

CRITICAL — scope rule:
  Answer ONLY what the agent literally asked, in the metric / scoring /
  submission-format domain. Do NOT volunteer extra information.

Allowed (only when the question is directly about it):
  - The exact name and a brief description of the evaluation metric.
  - Whether higher or lower scores are better.
  - The required submission column structure and value type
    (probability vs. hard label, score per class, top-K list, continuous
    regression value, etc.).
  - Any metric-specific quirks (e.g. K=5 for MAP@K; only certain target
    columns are scored; scoring is on inspiratory phase only; eps clipping;
    asymmetry of RMSLE).
  - Confirming or correcting an explicit assumption the agent stated.

Forbidden (treat ANY of these as out-of-scope, even if the agent frames
the question around the metric or asks you to "confirm"):
  - Do not reveal any test-set labels or test-set statistics.
  - Do not give exact CV scores, thresholds, or recommend models / toolchains
    / hyperparameters / training procedures / validation strategies.
  - Do not write code for the agent.
  - Do not copy the task description verbatim.
  - Do not mention internal benchmark details (manifest files, ground-truth
    paths, "decoy", etc.).
  - Do not advise on FEATURE-SET decisions: which input columns to use
    or drop (e.g. "use only the `text` field", "ignore `id`", "include
    column X"), even if the agent asks you to confirm.
  - Do not advise on FEATURE ENCODING: label-encoding, one-hot, target
    encoding, ordinal vs categorical, embeddings, scaling, etc.
  - Do not state or confirm rules about EXTERNAL DATA / pre-trained
    models / outside resources (e.g. "no external resources", "only local
    code", "no pretrained models"). The metric manifest does not cover
    competition rules.
  - Do not advise on TRAIN/VAL split, cross-validation, ensembling, or
    any modelling workflow detail.

If the agent asks you to do the modelling, write code, choose a toolchain,
suggest hyperparameters, choose features/encodings, confirm external-data
rules, or anything outside the metric / scoring / submission-format scope,
reply with EXACTLY this single line and nothing else:
"REFUSE: that's outside the scope of one clarifying answer."

Do NOT append any extra sentence. Do NOT volunteer the metric, the
submission format, the scoring direction, or any other clarifying
information when refusing. The refusal must be the entire reply.

If the agent's question contains BOTH an off-topic part AND an explicit
in-scope part (e.g. it asks both about hyperparameters AND about the
metric), answer ONLY the in-scope part in 1-2 sentences and ignore the
off-topic part — do not say "REFUSE" in this case, and do not mention
that you are ignoring anything.

If the agent's question is empty or just "NONE", reply: "NONE".

VARIANT NOTE: the agent's evaluation/metric section WAS
neutralised. In-scope metric/scoring/submission-format questions
are legitimate; answer them within the rules above.
\end{lstlisting}

\paragraph{Out-of-scope refusal template.}
This is the fixed response used when a clarification question asks for information outside the introduced ambiguity, such as test-set statistics, validation results, implementation advice, or construction details.

\begin{lstlisting}[style=promptstyle]
REFUSE: that's outside the scope of one clarifying answer.
\end{lstlisting}

\paragraph{Conservative ask-policy ablation.}
For the ask-policy sensitivity experiment, we also evaluate a stricter prompting variant that instructs the agent to ask only when the task cannot be resolved from the prompt and data package.

\begin{lstlisting}[style=promptstyle]
[CLARIFY PROTOCOL — read carefully]
Ask only if the task cannot be solved from the prompt and data. If the
task is fully specified or the answer can be inferred from local evidence,
write NONE. Unnecessary clarification is penalized.

  - You MAY first do a quick inspection of the data to inform your decision
    — e.g. read column headers, peek at a few rows, check basic shapes /
    dtypes, look at the sample submission. Keep this brief.
  - Then write your question (or the literal word `NONE`) to:
        <workspace>/_question.txt
  - Keep the question under 50 words and self-contained (the user only sees
    that file, not your tool calls or notes).
  - After writing the file, STOP. Do NOT train a model, run cross-validation,
    or write a submission in this turn. The user will reply in the next turn
    with the answer (or with `NONE`).
\end{lstlisting}

\subsection{Benchmark Construction Prompts}
\label{app:construction_prompts}

This subsection gives the prompts used to construct and verify Ambig-DS variants. The construction prompts make minimal edits that remove the relevant framing cue while preserving unrelated task information. The verification prompt checks plausible alternatives, ambiguity preservation, decision relevance, and task preservation.

\paragraph{Shared construction system prompt.}
Used for both target-ambiguity and objective-ambiguity construction.

\begin{lstlisting}[style=promptstyle]
You are an expert editor of data-science benchmark tasks.

You will be given an original task description and, when available, data/package information for reference. Your goal is to produce a minimally edited task description that removes one specific framing cue while preserving all unrelated task information.

Core principles:
1. Minimal edits. Change only the wording needed to remove the targeted framing cue. Do not rewrite unrelated sections.
2. Task preservation. The edited prompt must remain a natural, coherent data-science task description. It should not look corrupted, truncated, or intentionally adversarial.
3. Axis isolation. Remove only the targeted information. If editing target ambiguity, preserve the evaluation objective whenever possible. If editing objective ambiguity, preserve the target, data description, and submission format.
4. Ambiguity preservation. The edited prompt should not introduce new wording that reveals the hidden intended framing. The final prompt should remain compatible with at least two plausible framings along the targeted axis.
5. No new information. Do not add facts, assumptions, hints, examples, or constraints not present in the original task.

Return a JSON object:
{
  "text": "edited task description",
  "removed": ["specific phrases, sentences, or cues removed or changed"],
  "plausible_alternatives": ["at least two plausible framings"],
  "notes": "brief notes on why the edit preserves the rest of the task"
}
\end{lstlisting}

\paragraph{Target-ambiguity prompt rewriting.}
Used for DSBench target-ambiguity construction. The data package has already been transformed to contain two candidate targets, e.g., \texttt{val\_1} and \texttt{val\_2}; the prompt editor must not reveal which one is intended.

\begin{lstlisting}[style=promptstyle]
Edit the task description to remove target-identifying language while preserving the rest of the task.

The data package has been transformed so that the original target and a plausible decoy target appear as generic candidate columns, e.g., val_1 and val_2. The test set contains neither candidate target, and the sample submission uses a generic prediction column.

Your edit should make the intended prediction target underdetermined from the prompt.

Instructions:
- Remove or genericize sentences that name the target concept or target column.
- Replace target-specific phrases such as "predict count", "forecast demand", or "the target is sale price" with generic language such as "predict the value" or "produce predictions".
- Genericize mathematical variable definitions that reveal the target identity, e.g., replace "$p_i$ is the predicted count" with "$p_i$ is the predicted value".
- Remove target-identifying rows or headers in prompt examples, if they reveal the intended target.
- Remove or genericize data-field descriptions that reveal the original target or related target decomposition.
- Preserve unrelated task context, evaluation metric descriptions, file descriptions, and modeling instructions.
- Do not mention val_1, val_2, the decoy construction, or the hidden intended target.
- Do not add any hint that one candidate target is more likely than another.

Return the edited task description and metadata in the required JSON format.
\end{lstlisting}

\paragraph{Objective-ambiguity prompt rewriting.}
Used for MLE-bench objective-ambiguity construction. The target definition, data description, files, and required submission format are preserved.

\begin{lstlisting}[style=promptstyle]
Edit the task description to remove explicit evaluation-objective information while preserving the target and submission requirements.

Your edit should make the evaluation objective underdetermined from the prompt. The agent should still know what to predict and what file format to submit, but should not know which metric, scoring rule, optimization direction, thresholding rule, ranking rule, or metric-specific transformation will be used.

Instructions:
- Remove specific metric names such as AUC, log loss, RMSE, RMSLE, MAE, accuracy, F1, MAP@K, Pearson correlation, Levenshtein rate, or custom metric names.
- Remove metric formulas, optimization direction, leaderboard scoring rules, and metric-specific descriptions.
- Replace the Evaluation section with neutral language, e.g., "Submissions are evaluated against held-out ground truth using a standard evaluation procedure for this task."
- Preserve the target definition.
- Preserve the submission file columns and required file format.
- Preserve data descriptions, file lists, dataset context, and task background unless they explicitly reveal the metric.
- Do not make the prompt look corrupted or intentionally vague.
- Do not introduce a new metric or imply a particular metric through replacement wording.

Return the edited task description and metadata in the required JSON format.
\end{lstlisting}

\paragraph{LLM verification prompt (objective ambiguity).}
Used to verify candidate ambiguous-metric variants on MLE-bench before human review.

\begin{lstlisting}[style=promptstyle]
You are an expert reviewer for a benchmark on metric ambiguity in ML
competitions. You receive (a) the original Kaggle competition
description (full.md), (b) a redacted version (ambig_metric.md) that
must hide the evaluation metric, and (c) the structured manifest entry
recording the true metric.

Apply the four-item retention checklist from the benchmark paper:

  1. PLAUSIBLE ALTERNATIVES -- given only ambig_metric.md and the data
     package implied by it (column names, sample submission, etc.), at
     least two reasonable evaluation metrics remain consistent with the
     prompt. List them.

  2. AMBIGUITY PRESERVED -- the ambiguous variant does NOT leak the true
     metric anywhere: not in the Evaluation section, not in inline
     mentions ("predict a probability", "minimize ...", explicit metric
     names), and not in the submission-format hints. Cue leaks include
     formula fragments, optimization-direction wording, probability/
     hard-label hints that uniquely identify the metric, and
     metric-specific column semantics. Submission column NAMES are
     allowed to remain (the grader needs them); their METRIC-IDENTIFYING
     SEMANTICS are not.

  3. DECISION RELEVANT -- resolving the ambiguity changes a task-level
     choice a competent solver should make: hard labels vs probabilities,
     optimization direction, threshold/ranking behavior, top-K, clipping,
     column-wise aggregation, or submission semantics.

  4. TASK PRESERVED -- the redaction removes only metric-related
     information. Data descriptions, file lists, column definitions,
     submission column names, timeline, prizes, and citation are kept
     intact (modulo neutralized metric phrasing).

Output STRICT JSON, no markdown fences, no commentary. Schema:

{
  "checks": {
    "plausible_alternatives": {
      "pass": true|false,
      "rationale": "<= 2 sentences",
      "alternatives": ["<metric 1>", "<metric 2>", ...]
    },
    "ambiguity_preserved": {
      "pass": true|false,
      "rationale": "<= 2 sentences",
      "leaked_cues": ["<verbatim quote 1>", ...]
    },
    "decision_relevant": {
      "pass": true|false,
      "rationale": "<= 2 sentences"
    },
    "task_preserved": {
      "pass": true|false,
      "rationale": "<= 2 sentences"
    }
  },
  "verdict": "pass" | "fail",
  "notes": "optional <= 2 sentences"
}

`verdict` is "pass" iff all four checks pass. Be strict on
ambiguity_preserved: any verbatim quote that names or formulaically
identifies the true metric is a leak.
\end{lstlisting}

\paragraph{LLM verification prompt (target ambiguity).}
Used to verify candidate ambiguous-target variants on DSBench before human review. The checklist is identical in spirit but enumerates an axis-specific leak surface (target concept names, synonyms, two-target signposting, decoy/manifest references, leftover semantic feature names, sample-submission mentions, and numeric facts contradicting the manifest).

\begin{lstlisting}[style=promptstyle]
You are an expert reviewer for a benchmark on TARGET ambiguity in tabular
ML competitions. You receive (a) the original DSBench task prompt
(full.txt), (b) a rewritten variant (task_ambig.txt) that must hide the
original target column and add a second candidate target so that two
plausible targets coexist on disk, and (c) a manifest entry recording
which column is the true target and the anonymized->original feature
name map.

Apply the four-item retention checklist from the benchmark paper:

  1. PLAUSIBLE ALTERNATIVES -- given only task_ambig.txt and the data
     package implied by it (anonymised features f_01..f_NN, plus two
     candidate target columns val_1 and val_2 in train.csv), at least
     two reasonable target columns remain consistent with the prompt.
     List them.

  2. AMBIGUITY PRESERVED -- the ambiguous variant does NOT leak which
     column is the true target. Cue leaks include:
       * the original target concept name appearing in prose
         (e.g. "count", "price", "label", "target");
       * any synonym uniquely identifying the predicted concept
         (e.g. "rentals per hour", "house value");
       * explicit signposting that two candidate targets exist
         (the prompt must read as a single-target task);
       * mentions of "val_1", "val_2", "decoy", "ambiguous",
         "candidate target", "identify the target";
       * leftover semantic feature names from the manifest's
         feature_map (the prompt must use only f_01..f_NN);
       * sample_submission.csv being mentioned (it must be absent
         from the Files list);
       * numeric facts (n_train, n_test, n_features) that contradict
         the manifest -- only the manifest values are allowed.

  3. DECISION RELEVANT -- resolving the ambiguity changes a task-level
     choice a competent solver should make: which column is fit, which
     column appears in the submission file, and consequently which
     metric value the evaluator computes. (For target ambiguity this is
     true essentially by construction; rate fail only if the prompt
     somehow makes the choice trivial or moot.)

  4. TASK PRESERVED -- the rewrite removes only target-identifying and
     feature-semantic information. Dataset narrative, evaluation metric
     description, submission-format shape (header order, row ordering,
     id column), and section headers are kept intact (modulo target
     anonymisation and `prediction` as the generic header column).

Output STRICT JSON, no markdown fences, no commentary. Schema:

  {
    "checks": {
      "plausible_alternatives": {
        "pass": true|false,
        "rationale": "<= 2 sentences",
        "alternatives": ["<col 1>", "<col 2>", ...]
      },
      "ambiguity_preserved": {
        "pass": true|false,
        "rationale": "<= 2 sentences",
        "leaked_cues": ["<verbatim quote 1>", ...]
      },
      "decision_relevant": {
        "pass": true|false,
        "rationale": "<= 2 sentences"
      },
      "task_preserved": {
        "pass": true|false,
        "rationale": "<= 2 sentences"
      }
    },
    "verdict": "pass" | "fail",
    "notes": "optional <= 2 sentences"
  }

`verdict` is "pass" iff all four checks pass. Be strict on
ambiguity_preserved: any verbatim quote that names the original target
concept, names an original feature, signposts the two-target setup, or
contradicts the manifest counts is a leak.
\end{lstlisting}

\subsection{Framing Diagnostic Annotation Prompt}
\label{app:diagnostic_prompt}

\paragraph{Objective-framing judge prompt.}
For objective-ambiguity runs, we use an LLM judge to classify the agent's behavior from its code, logs, final response, and submitted predictions. The judge labels each run as aligned with the intended objective, aligned with a plausible alternative objective, degenerate, invalid/no-score, or other. We use majority vote over five judge calls, followed by human verification.

\begin{lstlisting}[style=promptstyle]
You read a data-science agent's full trajectory and final submission, and put it into ONE of five buckets describing what the agent did with respect to the competition's TRUE metric (given in the manifest).

The agent was given an ambiguous prompt that omits or paraphrases the metric. Return STRICT JSON. Quote short evidence verbatim from the trajectory.

SCHEMA (return exactly this JSON, no preamble, no code fences):

{
  "label":      "Intended" | "FormBroken" | "WrongObjective" | "Abdicated" | "Invalid" | "Other",
  "confidence": <float 0..1>,
  "evidence_quotes": ["<verbatim short snippet>", "..."],
  "rationale":  "<<= 2 sentences>"
}

BUCKETS (pick exactly one):

- "Intended": Agent built a real, non-degenerate model or heuristic whose
  optimization target is aligned with the TRUE metric (the same metric, a
  positively monotonic surrogate, or a hand-coded heuristic that approximates
  it), AND the submission file has the correct form (probabilities for
  AUC/log-loss; hard labels for accuracy/F1; text spans for Jaccard; reals for
  RMSE), AND the predictions actually use the input features (not constant or
  just the class marginals).

- "WrongObjective": Agent built a real model but optimized something not
  aligned with the TRUE metric (e.g. trained F1 instead of AUC with no
  monotonic relation; minimized when it should have maximized). Submission has
  real per-row signal but the optimization target is wrong.

- "Abdicated": Submission file exists but is degenerate / not a real model:
  copied `data/sample_submission.csv` or a provided baseline (e.g.
  `randomPredictions.csv`); every row predicts the same constant or the train
  marginals; trivial dummy like "first 30 chars of the context"; no real
  modeling step. Look for `cp data/sample_submission.csv` or
  `cp .../randomPredictions.csv` in the trajectory, or near-identical rows in
  the submission head.

- "Invalid": No usable submission. Either no _submission.csv at all, agent
  timed out before writing one, file is corrupt / unreadable / wrong schema
  the grader rejects, or trajectory is empty.

- "Other": Use ONLY if none of the five buckets above clearly applies (e.g.
  the run is a genuinely ambiguous edge case the schema doesn't cover). Prefer
  one of the five named buckets whenever possible.

DECISION ORDER (apply top to bottom, stop at first match):
  1. No submission / unreadable / agent timed out with no tools called -> Invalid.
  2. Submission is a copied baseline file, or every row is the same constant,
     or the agent narrates that it copied a baseline -> Abdicated.
  3. Agent trained an aligned objective but submission form contradicts the
     TRUE metric (e.g. AUC task with 0/1 labels) -> FormBroken.
  4. Agent built a real model with a non-aligned optimization target ->
     WrongObjective.
  5. Agent built a real, aligned model with correct submission form -> Intended.
  6. None of the above clearly applies -> Other.
\end{lstlisting}

\newpage

\end{document}